\documentclass{article}
\usepackage[utf8]{inputenc}

\usepackage{pgfplots} 

\pgfplotsset{compat=1.18}

\usepackage[pdfencoding=auto]{hyperref}

\usepackage{bm}

\usepackage{geometry}
\usepackage{graphicx}

\usepackage[citestyle=authoryear, backend=bibtex, sorting=nyt, natbib=true, labelnumber=true, url=false, doi=false]{biblatex}
\usepackage{amsmath}
\usepackage{amssymb} 
\usepackage{nicefrac}  
\usepackage{float} 
\usepackage{amsthm} 
\newtheorem{proposition}{Proposition}[section]

\usepackage{hyperref}

\usepackage{xcolor} 

\usepackage{thmtools} 

\declaretheorem[name=Hypothesis, preheadhook={}]{hyp} 

\usepackage{cleveref} 
\crefname{hyp}{hypothesis}{hypotheses} 
\Crefname{hyp}{Hypothesis}{Hypotheses} 

\usepackage{ragged2e} 

\usepackage{booktabs} 
\usepackage{multirow} 

\usepackage{algorithm}
\usepackage{algpseudocode}

\algnewcommand{\IfThen}[2]{
	\State \algorithmicif\ #1\ \algorithmicthen\ #2}
\algnewcommand{\LineComment}[1]{\State \(\triangleright\) #1}

\algnewcommand{\InlineFor}[2]{
	\State \algorithmicfor\ #1\ \algorithmicendfor\ #2}

 \makeatletter
\AfterEndEnvironment{algorithm}{\let\@algcomment\relax}
\AtEndEnvironment{algorithm}{\kern2pt\hrule\relax\vskip3pt\@algcomment}
\let\@algcomment\relax
\newcommand\algcomment[1]{\def\@algcomment{\footnotesize#1}}

\renewcommand\fs@ruled{\def\@fs@cfont{\bfseries}\let\@fs@capt\floatc@ruled
  \def\@fs@pre{\hrule height.8pt depth0pt \kern2pt}%
  \def\@fs@post{}%
  \def\@fs@mid{\kern2pt\hrule\kern2pt}%
  \let\@fs@iftopcapt\iftrue}
\makeatother

\usepackage{booktabs}
\usepackage{makecell}
\usepackage{pdflscape}

\usepackage[parfill]{parskip} 

\usepackage{authblk}

\usepackage{algorithm}
\usepackage{algpseudocode}

\providecommand{\keywords}[1]
{
  \small	
  \textbf{\textit{Keywords---}} #1
}

\title{Edge-Selector Model Applied for Local Search Neighborhood\\for Solving Vehicle Routing Problems}

\author[1]{Bachtiar Herdianto}
\author[1]{Romain Billot}
\author[2]{Flavien Lucas}
\author[3]{Marc Sevaux}
\author[4]{Daniele Vigo}

\date{\color{blue} 
      \nolinkurl{bachtiar.herdianto@imt-atlantique.fr}$^{*1}$\\
      \nolinkurl{romain.billot@imt-atlantique.fr}$^1$\\
      \nolinkurl{flavien.lucas@imt-nord-europe.fr}$^2$\\
      \nolinkurl{marc.sevaux@univ-ubs.fr}$^3$\\
      \nolinkurl{daniele.vigo@unibo.it}$^4$}

\affil[1]{IMT Atlantique, Lab-STICC (UMR 6285, CNRS), Brest, France}
\affil[2]{IMT Nord Europe, CERI Systèmes Numériques, Douai, France}
\affil[3]{Université Bretagne Sud, Lab-STICC (UMR 6285, CNRS), Lorient, France}
\affil[4]{DEI "G. Marconi", University of Bologna, Bologna, Italy}

\begin{document}

\maketitle

\begin{abstract}
    This research proposes a hybrid Machine Learning and metaheuristic mechanism that is designed to solve Vehicle Routing Problems (VRPs). The main of our method is an edge solution selector model, which classifies solution edges to identify prohibited moves during the local search, hence guiding the search process within metaheuristic baselines. Two learning-based mechanisms are used to develop the edge selector: a simple tabular binary classifier and a Graph Neural Network (GNN). The tabular classifier employs Gradient Boosting Trees and Feedforward Neural Network as the baseline algorithms. Adjustments to the decision threshold are also applied to handle the class imbalance in the problem instance. An alternative mechanism employs the GNN to utilize graph structure for direct solution edge prediction, with the objective of guiding local search by predicting prohibited moves. These hybrid mechanisms are then applied in state-fo-the-art metaheuristic baselines. Our method demonstrates both scalability and generalizability, achieving performance improvements across different baseline metaheuristics, various problem sizes and variants, including the Capacitated Vehicle Routing Problem (CVRP) and CVRP with Time Windows (CVRPTW). Experimental evaluations on benchmark datasets up to $30,000$ customer nodes, supported by pair-wise statistical analysis, verify the observed improvements.
\end{abstract}

\keywords{Metaheuristic, Vehicle Routing Problems, Binary Classification, Graph Neural Network}

\section{Introduction}
\label{sec:introduction}
    The Vehicle Routing Problems (VRPs) represent a class of combinatorial problems with many applications \citep{laporte2009fifty}. Solving them is challenging due to their $\mathcal{NP}$-hard nature, necessitating using effective search algorithms \citep{toth2014vehicle}. Among them, the Capacitated Vehicle Routing Problem (CVRP) is one of the most extensively studied \citep{toth2014vehicle,ALTABEEB2019105728,arnold2019knowledge,REZAEI2023110309}. Despite extensive research, the CVRP still presents a significant challenge in academic and industrial applications. \citep{prins2004simple,MARINAKIS2015680,YANG2025112988}. Recently, there has been increasing interest in leveraging ML to improve optimization algorithms \citep{kool2018attention,NEURIPS2020_f231f210,Hottung_Andr__2020,bengio2021machine,ma2021learning,morabit2021machine,Xin_Song_Cao_Zhang_2021,xin2021neurolkh,mesa2022machine}. The integration of ML with optimization algorithms can be categorized into three strategies: (1) end-to-end learning, (2) learning based on problem properties, and (3) learning repeated decisions \citep{bengio2021machine}. 
    
    The concept of learning from repeated decisions can be implemented by developing an in-loop ML-assisted optimization algorithm, enabling the baseline optimization algorithm to learn from its own decisions and adjust its behavior accordingly \citep{bengio2021machine}. In this paper, we introduce a hybrid optimization framework combining Machine Learning (ML) with metaheuristics to address VRPs. Our key contribution in this paper is the edge selector that identifies prohibited (tabu) moves during local search. We develop two learning-based mechanisms for the edge selector, that is the tabular binary classifier and GNN. Where for the binary classifier we utilize Gradient Boosting Trees (GBT) and Feedforward Neural Network (FNN) as baseline algorithm, with decision threshold optimization to address class imbalance. We further introduce a graph embedding pipeline that directly predict solution edges via structural pattern recognition. The proposed hybrid framework is integrated into two baseline metaheuristics: Fast Iterated Local Search with Localized Optimization (FILO) and Hybrid Genetic Search (HGS). It demonstrates improved performance across both CVRP and CVRPTW instances without requiring fine-tuning for each problem variant.
    
    \subsection{Vehicle Routing Problems (VRPs)}
    \label{subsec:vrps}
        In metaheuristic, local search operators are its important components, where it is proven as effective tool for improving solution. The initial idea of local search is that high-quality solutions to an optimization problem might be found by iteratively improving a solution using local neighborhood modifications \citep{toth2014vehicle}. Meanwhile, the utilization of ML, in particular the end-to-end learning framework by means of neural networks for optimization problems can be traced back to a Hopfield network that used to solve small instances of the Traveling Salesperson Problem (TSP) \citep{hopfield1985neural}. Tabu learning \citep{beyer1991tabu}, an extension of Hopfield neural networks, incorporates a dynamic, memory-based penalty term into the network’s energy function, preventing the network from revisiting previously encountered suboptimal solutions in continuous non-convex optimization problems. 
        
        Meanwhile, the capability of learning by processing graph data is highly important in combinatorial optimization, as many problems can be formulated as graphs \citep{bengio2021machine}. To this end, subsequent research introduced the graph ConvNets \citep{bresson2018residualgatedgraphconvnets} to learn rich representations of graph structures, producing edge probabilities that are then transformed into feasible TSP tours using beam search \citep{joshi2019efficient}. Building on this idea, the resulting heatmaps can also be leveraged to guide Dynamic Programming \citep{kool2022deep} or even local neighborhood search operators \citep{hudson2022graph}. A further advancement comes from NeuroLKH \citep{xin2021neurolkh}, which combines the graph ConvNet \citep{joshi2019efficient} with the LKH \citep{helsgaun2017extension}. In this framework, the graph ConvNet is used to assign edge scores and node penalties, which in turn guide the LKH.
        
        However, scalability and generalizability still need to be further investigated. In particular, prior research often requires fine-tuning every variants of the problem. Additionally, the quality of the solutions still needs improvement, as it remains lower than that achieved by already established methods. The learning repeated decision then falls into two approaches: optimization enhanced by learning and learning enhanced by optimization.
        
        \paragraph{\textbf{Learning Enhanced by Optimization}}
            In the first category, the ML model plays the primary role in refining solutions, while a simple search algorithm supports the process. These methods typically involve ML models that directly construct or guide the solution path. Early examples include NeuRewriter \citep{chen2019learning}, which used basic local search heuristics and Learn to Improve (L2I) \citep{Lu2020A}, which uses a reinforcement learning controller to select improvement operators based on solution states. Subsequent research has then focused on controlling the $k$-opt heuristic. One of the earliest works proposed a deep reinforcement learning to guide 2-opt \citep{wu2021learning}, later improved by the Dual-Aspect Collaborative Transformer (DACT) \citep{ma2021learning}, which introduced Dual-Aspect Collaborative Attention (DAC-Att) and a cyclic positional encoding method to guide 2-opt. Additionally, an RNN-based policy was proposed to govern 2-opt operations \citep{d2020learning}, which was later extended to 3-opt \citep{sui2021learning} by incorporating the Feature-wise Linear Modulation Network (FiLM-Net) \citep{Gupta2020_d1e946f4}. FiLM-Net modulates current solution features with selected link features as conditioning information to output appropriate reconnection types. However, these neural $k$-opt solvers are still constrained by small and fixed values of $k$. Further improvements were made in NeuroLKH \citep{ma2023learning}, which directly utilizes LKH \citep{helsgaun2017extension} as the baseline solver to improve subproblems guided by GNN-predicted heatmaps, allowing for more flexible $k$-move improvements. 
            
            Nevertheless, while some of these methods demonstrate generalizability, they still require fine-tuning for different problem variants. Moreover, future research is needed to enhance scalability and to demonstrate that the solution quality produced by these mechanisms can match or exceed that of current state-of-the-art solvers.
        
        \paragraph{\textbf{Optimization Enhanced by Learning}}
            In the second category, advanced optimization algorithms plays the primary role in refining solutions, with ML models providing auxiliary guidance to enhance decision-making within the search process. A representative example is the \textit{learning-to-delegate} mechanism introduced for large-scale CVRP instances \citep{li2021learning}, which integrates ML with state-of-the-art metaheuristic algorithms such as LKH \citep{helsgaun2017extension} and HGS \citep{vidal2022hybrid} to improve scalability. Their approach uses a Transformer-based model \citep{kool2018attention} to select subregions for refinement via subsolvers, in which the metaheuristic algorithms. Extending this direction, DeepACO \citep{ye2023deepaco} combines ML with Ant Colony Optimization (ACO) \citep{dorigo2007ant}, using GNNs to predict heatmaps to guide the probabilistic search process. This was later advanced by the Generative Flow Ant Colony Sampler (GFACS) \citep{kim2025ant}, which used Generative Flow Networks (GFlowNets) \citep{bengio2021flow} to learn a reward-proportional sampling distribution for guiding both ant sampling and pheromone updates. 
            
            The GNN \citep{gori2005new,scarselli2008graph} have also been applied to column selection for guiding column generation procedure \citep{morabit2021machine}, reducing computational time without compromising solution quality. In the domain of hybrid mechanisms aimed at enhancing local search, GNN-GLS \citep{hudson2022graph} leverages a graph ConvNet \citep{joshi2019efficient} to predict regret values for each edge in the problem instance, informing Guided Local Search (GLS) \citep{arnold2019knowledge} about which edges to penalize. In contrast, NeuroLS \citep{falkner2022learning} employed a $k$-GNN \citep{morris2019weisfeiler} to control the local search procedure directly. Further exploration into learning-based local search includes the Learning to Guide Local Search (L2GLS) framework \citep{SULTANA2024107136}, which employs policy network with attention-based encoding is trained to dynamically select local search operators at each step, based on the current solution state and recent action history. Further, the subsequent research introduces a learning-guided segmentation framework for CVRP that uses global graph-level features via Graph Attention Networks (GAT) \citep{veličković2018graph} to identify stable solution segments \citep{ouyang2025learningsegmentvehiclerouting}. These stable segments are aggregated and fixed, enabling the baseline heuristic to focus subsequent search iterations on the remaining unstable parts of the solution.
            
            Although these hybrid mechanisms show promise and in some cases outperform their respective baselines, much of the existing research lacks rigorous statistical validation to confirm significant improvements. Moreover, while many of these approaches are applicable across a range of routing problems, they typically require training or fine-tuning the ML models for each new problem variant, limiting generalizability.
        
    \subsection{Research Questions and Contributions}
    \label{subsec:research-question-and-contributions}
        The mechanism \textit{learning repeated decision} employs an in-loop ML-guided optimization framework, in which a baseline algorithm manages the high-level search process and iteratively invokes a machine learning model to guide lower-level decisions \citep{bengio2021machine}. This structure allows the algorithm to adapt its behavior over time by leveraging accumulated experience. 
        
        Although hybrid approaches combining ML and optimization have gained increasing attention \citep{Hottung_Andr__2020,bengio2021machine}, generalization remains a significant challenge \citep{bengio2021machine}, particularly in real-world applications where problem distributions can vary substantially. Another key concern is scalability, as hybrid models must be capable of handling large-scale and complex optimization problems \citep{bengio2021machine}. 
        
        Prior studies on VRP solvers indicate that solution-based features are often more effective for guiding optimization algorithms \citep{arnold2019knowledge,lucas2019comment}. At the same time, recent advances in Graph Neural Networks (GNNs) have demonstrated strong performance in diverse domains \citep{dwivedi2023benchmarking}, including their application in routing problems \citep{joshi2019efficient,kool2022deep,Joshi2022}. Building upon these insights, this research is designed to address the following questions:
    
        \begin{enumerate}
            \item How can an edge selector model be developed to classify solution edges and identify areas with potential improvement accurately?
            \item How can we design an effective hybrid mechanism that integrates the learned model to enhance the performance of the metaheuristic baselines?
            \item How can we ensure that the hybrid mechanism generalizes sufficiently across different problem variants while avoiding the need for fine-tuning the learning model?
        \end{enumerate}
    
        To address these questions, we propose the \textit{edge selector model}, a ML model capable of classifying the edges of a VRP solution. This model aims to dynamically control the neighborhood exploration during local search improvement phase while filtering unnecessary moves by reducing the search space. In summary, our contributions are:
    
        \begin{enumerate}
            \item The edge selector model that predicts areas of improvement in a VRP solution, filtering non-promising moves within local search neighborhoods.
            \item A hybrid mechanism that integrates the proposed edge-solution selector model with metaheuristic baselines to solve the CVRP.
            \item A generalized hybrid mechanism by extending the developed for the CVRP for solving the CVRPTW without fine-tuning the baseline ML model.
        \end{enumerate}

        The remainder of this paper is organized as follows: \Cref{sec:baseline-metaheuristics} presents the baseline metaheuristics used as baseline for developing our proposed hybrid mechanism. \Cref{sec:proposed-method} details the architecture and components of our proposed hybrid pipeline, while \Cref{sec:generalization} explains the generalization principle operated in the proposed hybrid mechanism. \Cref{sec:computational-experiment} shows the experimental setup and comparative evaluation of our proposed hybrid mechanism. Finally, \Cref{sec:conclusion} summarizes our findings.

\section{Baseline Metaheuristics}
\label{sec:baseline-metaheuristics}
    Local search neighborhoods form the important mechanism of metaheuristics \citep{Prodhon2016,arnold2019knowledge}. Metaheuristics for solving VRPs in general can be categorized into two main classes: \textit{path-based methods} and \textit{population-based methods} \citep{Prodhon2016}. Path-based methods employ a simple iterative mechanism during optimization, where local search neighborhoods are applied to progressively improve solution. These methods typically incorporate solution acceptance criteria to balance exploration and exploitation. A representative example is the Fast Iterated Local Search Localized Optimization (FILO) metaheuristic \citep{accorsi2021fast}, which consists of a construction phase, followed by an improvement phase involving the optional route minimization and core optimization procedure that applies various local search neighborhoods. Meanwhile, population-based methods maintain and evolve a diverse set of solutions simultaneously. These approaches utilize recombination operators coupled with local search mechanisms to guide the population toward high-quality feasible solutions. The Hybrid Genetic Search (HGS) metaheuristic \citep{vidal2022hybrid} represents this category, maintaining both feasible and infeasible solutions in its population while employing local search neighborhoods to drive the optimization process toward feasible good solutions.
    
\begin{figure}[htbp] 
    \centering
    \includegraphics[width=1\textwidth]{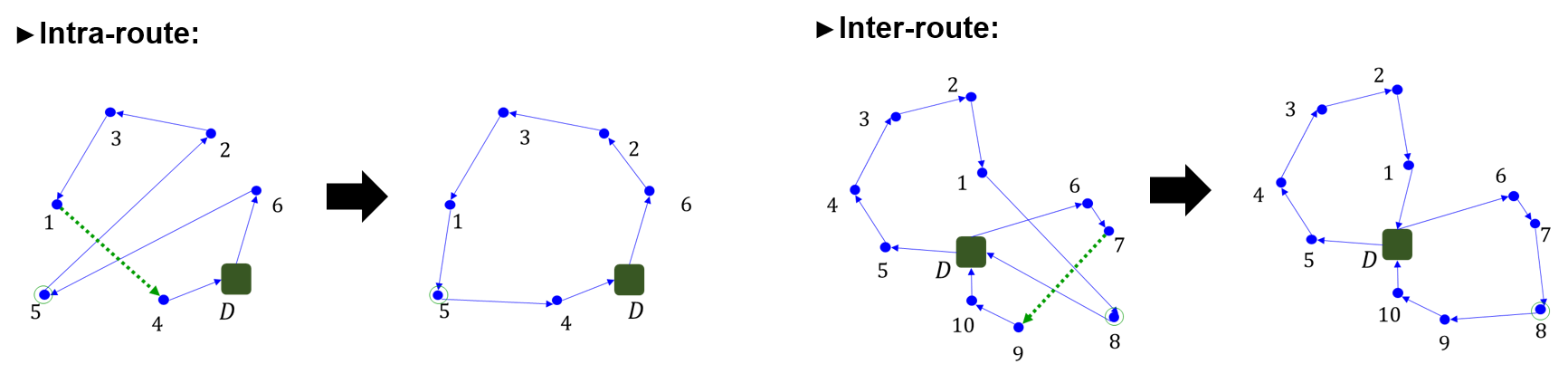}
    \caption{Reallocation involves relocating a node to a new position within the current solution.} \label{fig:ls-insert} 
\end{figure}
    
    \paragraph{\textbf{Local search neighborhoods}} 
        Local search provides a simple yet effective procedure for improving solution in combinatorial optimization in which tasks of the type $\min_{x \in X} c(x)$, where $X$ is the set of feasible solutions and $c$ denotes the cost function. The neighborhood $\mathcal{N}_{\tau}(x)$ of a solution $x$ is the set of all solutions that can be obtained by applying any move $\tau$ from a predefined set of transformations $\mathcal{T}$ to $x$, $\mathcal{N}_{\tau}(x) = \{ \tau(x) \mid \tau \in \mathcal{T} \} \subset X$. In the case of the CVRP, the neighborhood $\mathcal{N}_{\tau}(x)$ is commonly determined relative to a set of moves that can modify the current solution $\mathcal{S}$. Each move $\tau$ corresponds to a minor change that transforms $x$ into a neighboring solution $\tau(x) \in \mathcal{N}_{\tau}(x)$. \Cref{fig:ls-insert} illustrates examples of local search moves. The local search operation reallocates a customer node by inserting it at a new position in the current solution. This move can be performed within the same route (intra-move) or between different routes (inter-move).

    \paragraph{\textbf{Tabu search mechanism}}
        Tabu Search escapes local optima by keeping a tabu list of recent moves. A move is prohibited if it matches any entry in this list. Further, the concept of an aspiration criterion allows the algorithm to override the tabu status when a move results in a better solution, promoting flexibility in the search process \citep{glover1998tabu}.

\begin{figure}[htbp] 
    \centering
    \includegraphics[width=1\textwidth]{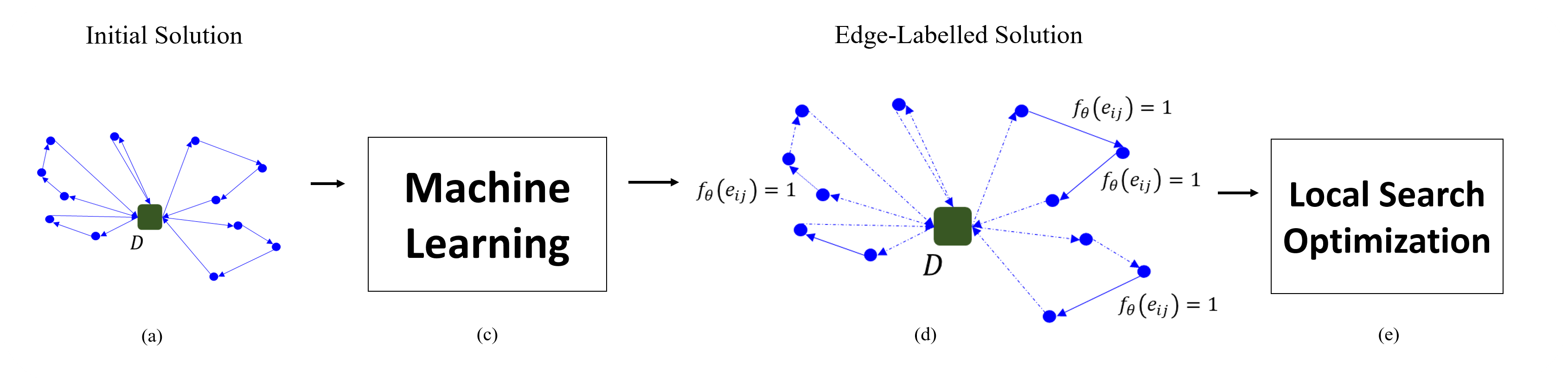}
    \caption{Overview of the proposed hybrid framework.} \label{fig:iterative-framework} 
\end{figure}

\section{The Edge Solution Selector Model}
\label{sec:proposed-method}
    In the proposed hybrid mechanism,  prohibited (tabu) moves are predicted before performing local search. This is accomplished by developing an edge selector model $f_{\theta}$ that assigns binary labels $f_{\theta}(e_{ij}) \in \{0, 1\}$ for every edge $\forall i,j \in E$ in the graph. By labeling these edges, we can control the behaviour of the local search, aiming to reduce the search space and avoid unnecessary moves. As illustrated in \Cref{fig:iterative-framework}, the mechanism begins by constructing the initial solution $S_0$ in (a) from the problem instance. The proposed edge selector model $f_{\theta}$ is implemented in (b), which predicts labels for all edges in $E(S_0)$ (c). These labels guide the improvement process, where further improvements are achieved via a set of local search operators (d). Moreover, when the metaheuristic baseline employs a multi-start search strategy \citep{Prodhon2016}, the process of construction $S_0$, prediction, and local search is repeated iteratively until the termination criterion is met.

\begin{figure}[htbp] 
    \centering
    \includegraphics[width=0.88\textwidth]{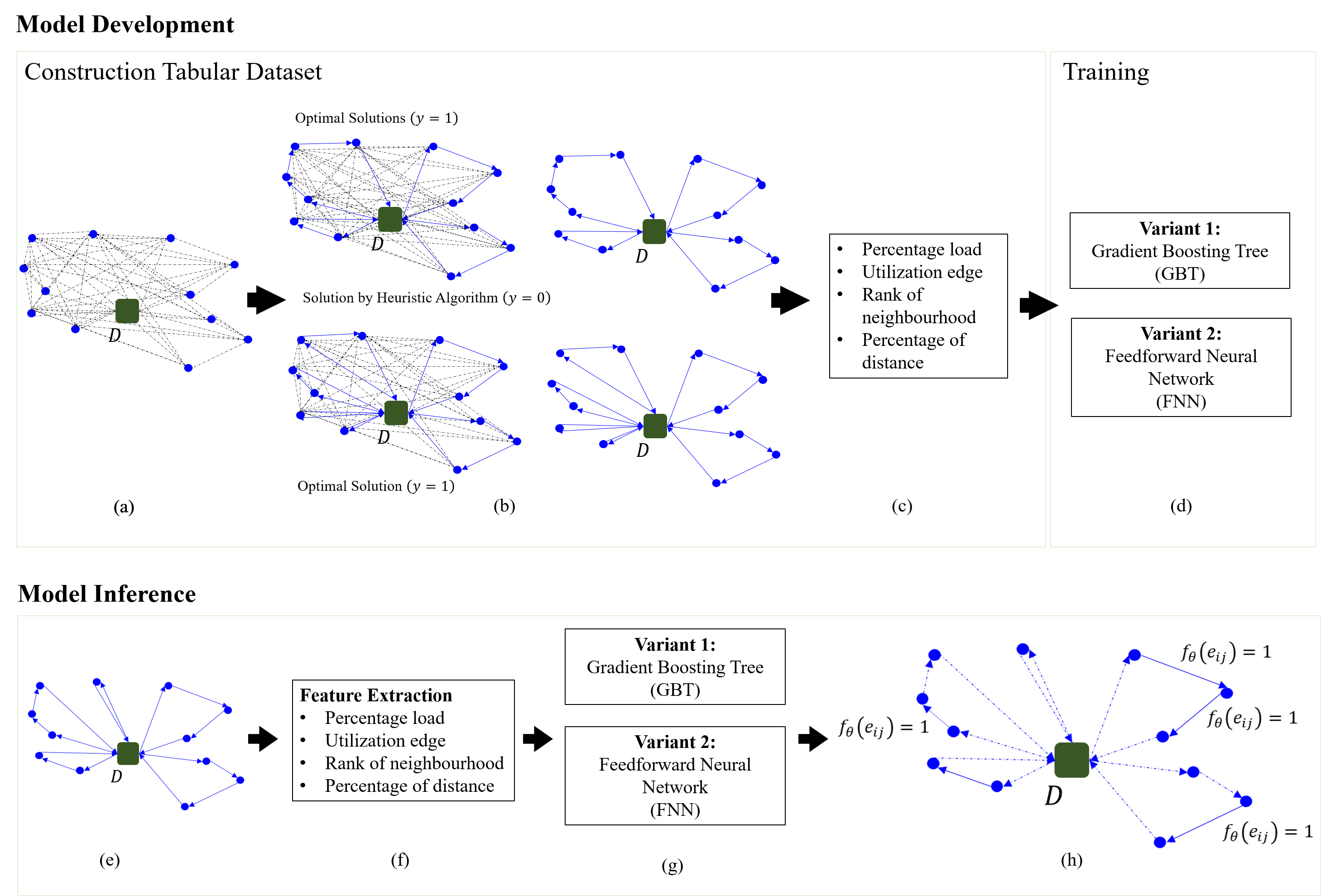}
    \caption{Overview of the hybrid framework using tabular classifiers as the baseline learning model.} \label{fig:detailed-tabular} 
\end{figure} 

    \subsection{Tabular Binary Classification}
    \label{subsec:ml-model}
        In this mechanism, the solution $S_0$ is transformed into a two-dimensional feature matrix $\mathbf{X}$. This matrix is constructed using $p$ features that describe the edge structure of $S_0$. Each edge in $S_0$, denoted as $E(S_0) \subseteq E$, is then labeled by the selector $f_{\theta}$. The resulting matrix $\mathbf{X}$ has dimensions $|E(S_0)| \times p$, where $|E(S_0)|$ is the number of edges in $S_0$ and $p$ is the number of features. The selector function $f_{\theta}$ operates as follows:

        \begin{equation} \label{eq:tabular-mechanism}
            f_{\theta}: E(S_0) \rightarrow \{0,1\}
        \end{equation}
        
        As illustrated in \Cref{fig:detailed-tabular}, the development of the tabular model involves two main stages: constructing the tabular solution-edge dataset for training (a to c), and training the baseline classifier through a supervised learning approach (d). The dataset construction begins by solving $10,000$ instances from the $\mathbb{XML}$100 benchmark \citep{queiroga202110}, which includes both optimal\footnote{Provided by \citet{queiroga202110}} and non-optimal solutions. As shown in subfigure (b), the non-optimal solutions are generated via the Savings ($S_{cw}$) and Sweep ($S_{sweep}$)\footnote{Implemented using the solver by \citet{groer2010library}: \url{https://github.com/coin-or/VRPH}} heuristics, as well as the MNS-TS ($S_{mns}$) \citep{soto2017multiple} algorithm. The resulting tabular dataset used to train the classifier baseline consists of pairs of edge features and labels $(x_{(i)}, y_{(i)})$ between customer nodes $C$, where $x_{(i)}, y_{(i)} \in \mathcal{D}$ and $i \in \mathbb{N}$. Further details about the features used in the tabular mechanism are provided in subfigure (c) and described in \ref{sec:app-feature}. As depicted in (d), we develop two classifier variants of the selector model $f_{\theta}$ for the tabular mechanism: Gradient Boosting Tree (GBT) and Feedforward Neural Network (FFN). Moreover, designing the tabular learning architecture $f_{\theta}$ involves defining the classifier structure and addressing class imbalance, which arises from the disparity between the number of edges in the solution and the total number of possible edge combinations. To ensure both models output a probability score indicating the likelihood of each edge belonging to the optimal solution, a sigmoid activation function $\sigma(z) = \nicefrac{1}{1+e^{-z}}$ is applied at the final layer. Lastly, as illustrated in \Cref{fig:detailed-tabular} (e to h), during inference with the hybrid mechanism using the tabular baseline, the process begins by extracting an initial solution, as shown in subfigure (e). Edge-solution features are computed in (f) and then passed to the classifier in (g), which produces an edge-labeled solution in (h). This labeled solution is subsequently used to guide the improvement phase, where further enhancements are achieved using a set of local search operators.

        \paragraph{\textbf{Gradient Boosting Tree}}
            In the first variant, the tabular variant of the selector $f_{\theta}$ is implemented using the GBT. Given an input vector $\mathbf{X} \in \mathbb{R}^p$, the GBT model with $\mathbb{M}$ estimators performs binary classification. Let $h_m(\mathbf{X})$ be the prediction from the $m$-th estimator, the final prediction is:
            \begin{equation} \label{eq:gbt}
                f(\mathbf{X}) = \sigma(F_M(\mathbf{X}))= \sigma \left( \sum_{m=1}^{\mathbb{M}} \gamma_m h_m(\mathbf{X}) \right)
            \end{equation}
            where $\gamma_m$ is its learning rate and $\sigma(\cdot)$ is the sigmoid activation function, mapping outputs to the range $(0,1)$.
        
        \paragraph{\textbf{Feedforward Neural Network}}
            In the second variant, we implement the tabular variant of the selector $f_{\theta}$ using the FNN as a baseline classifier. Given an input vector $x \in \mathbb{R}^p$, the FNN consists of $\mathbb{L}$ layers for binary classification. Let $M_l$ denote the number of neurons in the $l$-th layer, with $M_0 = p$, $M_{\mathbb{L}} = 1$, and $M_l = d_l$ for $0 < l < \mathbb{L}$. The weight matrix and bias vector for layer $l$ are denoted as $\mathbf{W}_l$ and $\mathbf{b}_l$, respectively. The selector's prediction is given by:
            \begin{equation} \label{eq:fnn} 
                f(\mathbf{X}) = \sigma \left(\mathbf{W}^{\top}_{\mathbb{L}} \dots \mathrm{ReLU} \left( \mathbf{W}^{\top}_1 x + \mathbf{b}_1 \right) \dots + \mathbf{b}_{\mathbb{L}} \right)
            \end{equation}
            Each hidden layer uses the ReLU (Rectified Linear Unit) as the activation function, $\mathrm{ReLU}(z) = \max(0, z)$, and the last layer applies the sigmoid function $\sigma(\cdot)$.
    
        \paragraph{\textbf{Thresholding under class imbalance}}
            As mentioned earlier, the main challenge in this mechanism is handling class imbalance, as the number of solution edges is much smaller than the total possible edges. To address this in tabular binary classification, we explore two thresholding strategies. The sigmoid output $\sigma$ from the baseline algorithms require a threshold $t$, which plays a critical role in imbalanced settings \citep{narasimhan2014statistical,singh2022optimal}. For the selector $f_{\theta}$, the goal is to minimize false positives while accurately identifying true prohibited moves. While the default threshold is $t = 0.5$, in this research, we apply a deterministic rule:
            \begin{equation} \label{eq:deterministic-th}  
                \hat{f}(x) = 
                \begin{cases} 
                    1 & \text{if } f(x) > t\\ 
                    0 & \text{otherwise}
                \end{cases}
                \quad \text{with } 0 < t < 1
            \end{equation}
            Alternatively, we also apply stochastic threshold decision rule to account robustness near the decision boundary.
            \begin{equation}\label{eq:stochastic-th}
                \hat{f}(x) = 
                \begin{cases} 
                    1 & \left( f(x) > t+\epsilon \right) \,\vee\, \left( \text{abs} \left(f(x)-t \right) < \epsilon \,\land\, p > \mathcal{R} \right)\\
                    0 & \text{otherwise} 
                \end{cases} 
            \end{equation}
            where $\epsilon$ is a small positive constant used as a tolerance for numerical comparison and $\mathcal{R} \sim \mathcal{U}(0,1)$. Both methods are applied to improve the performance of the selector $f_{\theta}$ under class imbalance.

\begin{figure}[htbp]
    \centering
    \includegraphics[width=0.88\textwidth]{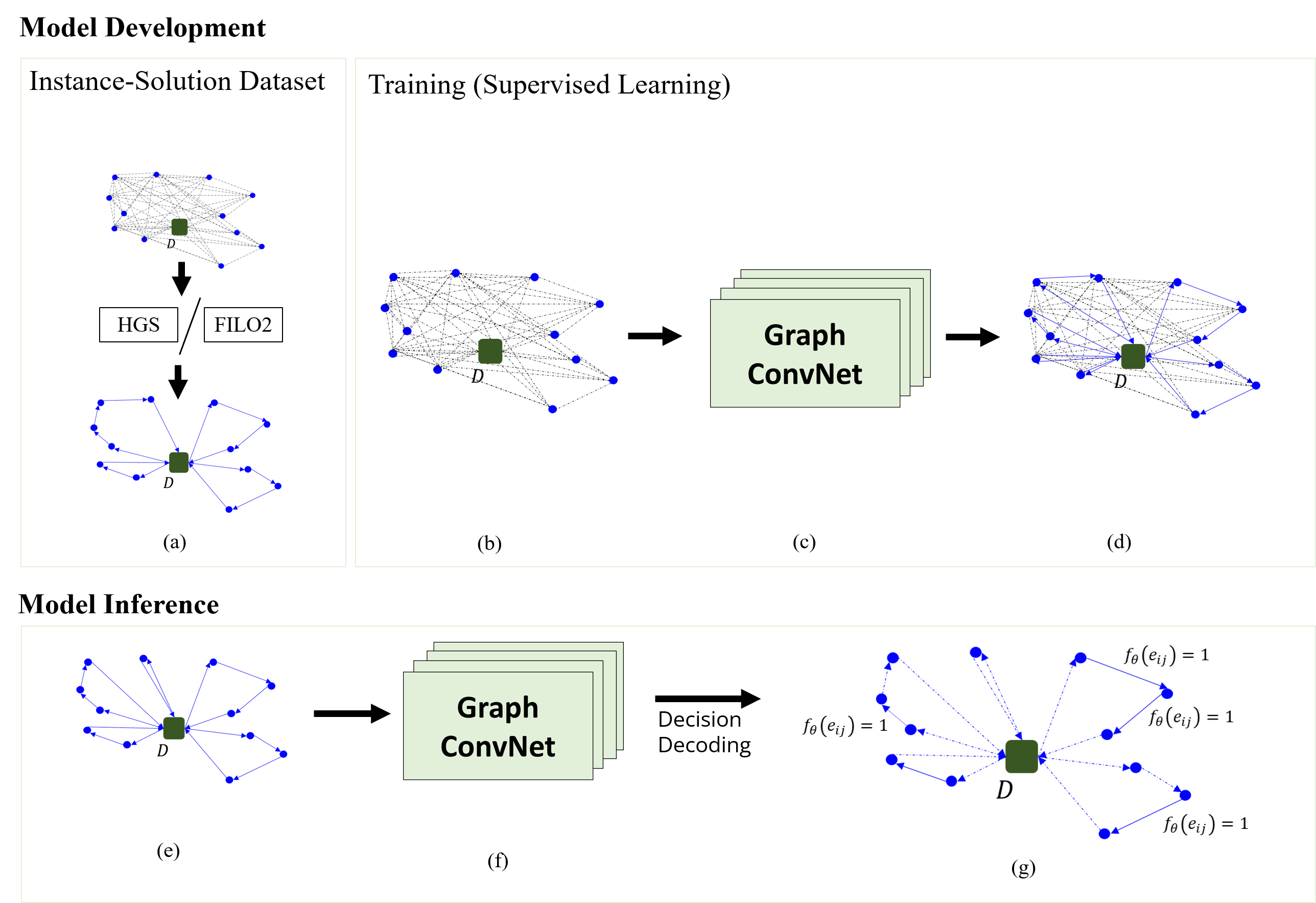}
    \caption{Overview of the hybrid framework with Graph ConvNet as the baseline learning model.} \label{fig:overview-gnn}
\end{figure} %

    \subsection{Graph Neural Networks}
    \label{subsec:graph-learning}
        Recent GNN-based hybrid optimization solvers incorporated GNNs with heuristic search to approximate solutions for graph-structured problems \citep{joshi2019efficient,Joshi2022,kool2022deep,dwivedi2023benchmarking}. Subsequently, to further enhance the performance of the selector $f_{\theta}$, we explore the use of GNNs as a baseline learning model. In this research, we build upon these existing methods by leveraging edge prediction capabilities in GNNs to guide the search process, in particular, to identify prohibited moves during local search neighborhood exploration. Similar to the tabular baseline, the development of the graph-based model, as illustrated in \Cref{fig:overview-gnn}, involves two main stages: (1) constructing the training dataset (a), and (2) training the model (b to d). The dataset, consisting of CVRP instances and their corresponding solutions, is divided into small-scale and large-scale instances. The baseline graph learning model is then trained using this dataset through a supervised learning approach. During inference, the hybrid mechanism begins by generating an initial solution $S_0$ (e), which is then passed to the edge selector model $f_{\theta}$ (f). The selector then predicts the likelihood of each edge in $S_0$ being part of a high-quality solution. This results in an edge-labeled solution (h), which is then used to guide the improvement phase. Here, a set of local search operators is applied in this phase to refine the solution further.
        
        In \Cref{fig:overview-gnn}, the initial solution $S_0$ is used as input to the selector $f_{\theta}$. The $S_0$ can be represented a sparse graph, which serves as the basis for predicting the likelihood that its edges belong to a high-quality solution. We extend the probabilistic output of the selector $f_{\theta}$ by applying a deterministic threshold $t$, as defined in \Cref{eq:deterministic-th}, to determine whether an edge should be prohibited from being involved in local search moves.

        \paragraph{\textbf{Graph embedding}}
            We extend the graph Convolutional Network (ConvNet) architecture \citep{joshi2019efficient,kool2022deep} to develop a hybrid node-destroyer model that guides the removal of customer nodes during local search. Unlike prior works that primarily employ graph ConvNets to construct solutions from scratch \citep{joshi2019efficient,kool2022deep,Joshi2022,dwivedi2023benchmarking}, our approach leverages the model to evaluate and select components within an existing solution. The graph ConvNet incorporates anisotropic message passing with edge gating, enabling the iterative update of both node and edge representations across network layers. Each input node feature is represented as a three-dimensional vector $x_i \in [0,1]^3$, encoding the node’s coordinates and its demand utilization $q/Q$.

\begin{figure}[htbp]
    \begin{center}
        \includegraphics[width=0.38\textwidth]{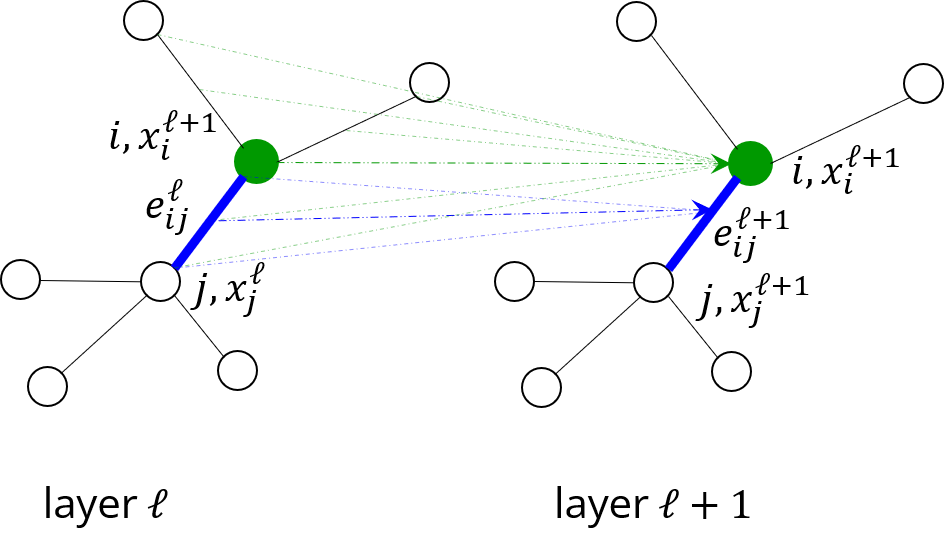}
        \caption{Representation message passing in graph ConvNet.}
        \label{img:edge-convnet}
    \end{center}
\end{figure}  

            The edge features, corresponding to the rounded Euclidean distances $d_{ij}$, are embedded into an $h/2$-dimensional vector space. Let $x_i^{\ell}$ and $e_{ij}^{\ell}$ denote the node and edge feature vectors at layer $\ell$, respectively. The node update rule is defined as:
            \begin{equation}\label{eq:node-feature}
                x_i^{\ell+1} = x_i^{\ell} + \mathrm{ReLU} \left( \mathrm{BN} \left( W_1^{\ell} x_i^{\ell} + \sum_{j \sim i} \eta_{ij}^{\ell} \odot W_2^{\ell} x_j^{\ell} \right) \right)
            \end{equation}
            where $\eta_{ij}^{\ell}$ is the attention map, computed as:
            \begin{equation}
                \eta_{ij}^{\ell} = \frac{\sigma\left( e_{ij}^{\ell} \right)}{\sum_{j' \sim i} \sigma\left( e_{ij'}^{\ell} \right) + \epsilon}
            \end{equation}
            with $\sigma(\cdot)$ denoting the sigmoid activation and $\epsilon$ a small constant for numerical stability. The edge features are updated using the following rule:
            \begin{equation}\label{eq:edge-feature}
                e_{ij}^{\ell+1} = e_{ij}^{\ell} + \mathrm{ReLU} \left( \mathrm{BN} \left( W_3^{\ell} e_{ij}^{\ell} + W_4^{\ell} x_i^{\ell} + W_5^{\ell} x_j^{\ell} \right) \right)
            \end{equation}
            
            Here, $W \in \mathbb{R}^{h \times h}$ denotes a learnable weight matrix. The function $\mathrm{BN}$ refers to Batch Normalization, which normalizes activations within each mini-batch. As illustrated in \Cref{img:edge-convnet}, the model stacks multiple layers to refine node and edge embeddings iteratively. The green and blue arrows in the figure indicate the information flow between nodes and edges, enabling the model to learn structural representations from the input graph to estimate the likelihood that each edge belongs to a high-quality solution.

        \paragraph{\textbf{Curriculum learning}}
            We initially train our GNN-based selector $f_{\theta}$ on the CVRP instances and solution dataset of size $N = 100$ \citep{kool2022deep}. Hence, to improve the scalability and generalization of the baseline model to handling larger problem sizes, we leverage a curriculum learning strategy, where the model is progressively trained on instances of increasing complexity \citep{bengio2009curriculum}. In particular, the selector $f_{\theta}$ is further trained on larger instance-solution dataset, with size $N = 1,000$. This dataset is generated using the $\mathbb{XML}$100 instance generator\footnote{Available at \url{http://vrp.galgos.inf.puc-rio.br/index.php/en/}} \citep{queiroga202110}. A total of $100,000$ instances are generated and then solved by using the FILO2 \citep{ACCORSI2024106562}\footnote{These instances and solutions are available at \url{https://github.com/bachtiarherdianto/MH-Edge-Selector}}, which provides high-quality labels for further training the model. This strategy enables the selector $f_{\theta}$ to scale on larger CVRP instances.

\begin{figure}[htbp]
    \centering
    \includegraphics[width=1\textwidth]{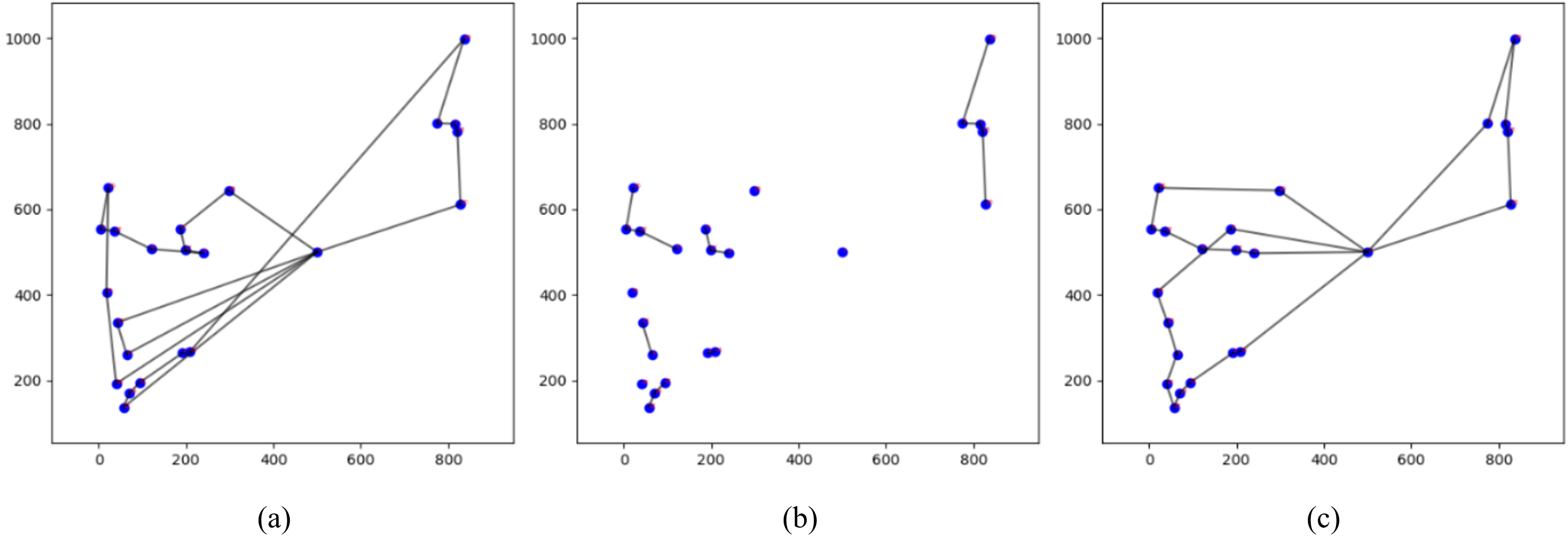}
    \caption{Overview model inference from a CVRP instance with $n=22$.} \label{fig:fixed-edges} 
\end{figure}

        \paragraph{\textbf{Model inference}}
            While the training phase utilizes the full graph for the CVRP instances with $N = 100$, and a sparse graph with $k = 25$ nearest neighbors for $N = 1,000$, the inference phase adopts a different approach. Specifically, during inference, edge evaluation is restricted to those forming the initial solution $E(S_0) \subseteq E$, generated by the baseline metaheuristic. In other words, the model operates on a sparse graph, derived from $S_0$, as illustrated in \Cref{fig:fixed-edges}. The performances of the selector $f_{\theta}$ during model inference can be measured using the precision score. The goal of $f_{\theta}$ is to maximize the number of edges correctly classified as prohibited moves that are indeed part of the high quality (groundtruth) solution, while minimizing the number of edges wrongly predicted as prohibited that do not belong to the groundtruth.  Let $\text{TP}$ (\textit{True Positive}) be defined as $\text{TP} := \mathbb{P} \left( \hat{f}(\mathbf{x}) = 1 \land y = 1 \right)$, and let $\text{FP}$ (\textit{False Positive}) be defined as $\text{FP} := \mathbb{P} \left( \hat{f}(\mathbf{x}) = 1 \land y = 0 \right)$. This aligns with the binary classification setting, in which maximizing true positives and minimizing false positives. Thus, the precision score is computed as:
            
            \begin{equation} \label{eq:precision}
                \text{precision} := \mathbb{P} \left( y = 1 \mid \hat{f}(\mathbf{x}) = 1 \right) = \dfrac{\mathbb{P} \left( \hat{f}(\mathbf{x})=1 \land y=1\right)}{\mathbb{P}(\hat{f}(\mathbf{x})=1)} = \dfrac{\text{TP}}{\text{TP}+\text{FP}} 
            \end{equation}

            As illustrated in \Cref{fig:fixed-edges}, for an instance with $22$ customer nodes, the initial solution $S_0$ is shown in subfigure (a). The selector predicted $26$ prohibited edges, as depicted in subfigure (b). When comparing subfigure (b) to the groundtruth in subfigure (c), we observe that $22$ of these edges are $\text{TP}$, while the remaining $4$ are $\text{FP}$, resulting in a precision score of $0.8461$. This demonstrates that the predictions made by the selector $f_{\theta}$ closely align with the ground truth structure.

            \paragraph{\textbf{Solving very large scale problems}} 
                Utilizing sparse graphs helps speed up the training and inference phases without a major loss in prediction quality \citep{kool2022deep}. For instances larger than $1,000$ customer nodes, the size of the full graph results in significant computational overhead. To manage this, inference is limited to the subset of edges that define the initial solution $S_0$. Accordingly, only the $1,000$ nearest nodes to the depot $c_0$ are used by the selector $f_{\theta}$ in very large-scale problem instances. Experimental analysis during model development confirms that this scalable strategy lowers runtime computation without sacrificing prediction quality.

\section{Generalization}
\label{sec:generalization}
    While previous research has explored generalization in constructive algorithms within end-to-end learning framework \citep{Hottung_Andr__2020,NEURIPS2023_a68120d2,NEURIPS2023_9c93b3cd}, our study presents a framework that generalizes through iterative learning repeated decision mechanism. Given that a feasible VRP solution consists of a set of routes $R$, each starting and ending at the depot $c_0$, where every route $r_k$ satisfies the capacity constraint $q(r_k) \leq Q$ and each customer is visited exactly once, we investigate two aspects of generalization. The first aspect is applying the selector $f_{\theta}$ to different VRP variants, such as CVRP and CVRPTW. And the last one is integrating the selector $f_{\theta}$ with different baseline metaheuristics. \Cref{prop:generalization} presents the generalization capability of the proposed hybrid mechanism, specifically in solving VRPs.

    \begin{proposition}\label{prop:generalization}
        The selector $f_{\theta}$, designed to fix (prohibited) specific edges rather than generate complete solutions, can generalize beyond the CVRP to other VRP variants.
    \end{proposition}

    \begin{proof}
        Let $\mathcal{F}$ denote the set of fixed (prohibited) edges in solution $S$, and $E$ the set of remaining (adjustable) edges, such that $\mathcal{F} \cup E = S$. Local search operators, when applied under the feasibility constraints of a problem, produce a modified set $E'$. Since both $\mathcal{F}$ and $E'$ satisfy the constraints of the VRPs, the new solution $S' = \mathcal{F} \cup E'$ remains feasible.
    \end{proof}

\section{Experiment and Analysis}
\label{sec:computational-experiment}
    We implemented the baseline solver in C++ and selector $f_{\theta}$ in Python\footnote{Available at \url{https://github.com/bachtiarherdianto/MH-Edge-Selector}}. The experiments were conducted on a 64-bit Debian GNU/Linux 12 machine with 16 virtual AMD cores, 62 GB RAM, and an NVIDIA L40 GPU. Considering the pseudo-random aspect \citep{matsumoto1998mersenne}, each experiment involved five runs, with the random seed was defined as the run counter minus one. Throughout the experimentation, we refer to the following:

    \begin{itemize}
        \item BKS: the total cost value of the best-known solutions. All the information related to the instances and best-known solutions are available at \url{http://vrp.galgos.inf.puc-rio.br/index.php/en/}.
        \item Gap: represents the relative deviation of a solution from the optimal solution, or from the best-known solution (BKS) if the optimal remains unknown. It is given by:
        \begin{equation}
            \text{Gap} = \dfrac{\text{Obtained Solution} - \text{BKS}}{\text{BKS}} \times 100\%
        \end{equation}
    \end{itemize}
    
    \paragraph{\textbf{Experiment setup}}
        The selector $f_{\theta}$ is developed using two learning-based mechanisms: the tabular binary classifier and a graph learning, as summarized in \Cref{table:variant-filo2-hybrid}. For the tabular models, the baseline GBT requires only around $8$ minutes for training, while the FNN takes approximately $20$ hours. However, both training processes are relatively short and can be performed without GPU acceleration, as the learning architecture for the tabular models is significantly simpler compared to the graph-based model. Subsequently, training the baseline graph-based model is more heavy. It was conducted supported with NVIDIA L40 GPU. During evaluation on the validation and test sets, the adjacency matrix, which represents the probabilities of all edges in the initial solution obtained from the graph ConvNet, is converted into binary decisions. 
        
        Using high-quality solutions as ground-truth labels, precision is computed as described in \Cref{subsec:graph-learning}. The selector model $f_{\theta}$ was first trained on CVRP instances with $N = 100$, using labels generated from high-quality HGS solutions \cite{kool2022deep}. This initial training phase took approximately $1.6$ hours and used about 2 GB of GPU memory. To adapt $f_{\theta}$ to more complex problems with $N = 1,000$, we employed a curriculum learning strategy by further training the model on larger instances. This curriculum learning phase consumed $42.9$ GB of GPU memory and lasted around $7.5$ hours. The curriculum learning approach improved the model's effectiveness on large-scale instances. 

\begin{table}[htbp]
\caption{Various Hybrid Mechanisms with Baseline FILO2.}
\label{table:variant-filo2-hybrid}
    \centering
    \scalebox{0.8}{
    \hspace*{1cm}
    \begin{tabular}{r c | c c c c c c}
        \toprule
            \multirow{3}{*}{\makecell{\textbf{Hybrid} \\ \textbf{Mechanism}}} && 
            \multicolumn{3}{c}{\textbf{Learning Model}} && 
            \multicolumn{2}{c}{\textbf{Binary Thresholding}} \\
            \cmidrule(lr){3-5} \cmidrule(lr){7-8}
            && \multicolumn{2}{c}{\textbf{Tabular}} & \textbf{Graph} 
            && \multirow{2}{*}{\textbf{Deterministic}} & \multirow{2}{*}{\textbf{Stochastic}} \\
            \cmidrule(lr){3-4} \cmidrule(lr){5-5}
            && GBT & FNN & ConvNet &&& \\
            \midrule
            FILO2-$\alpha$ && \checkmark & & && \checkmark & \\
            FILO2-$\beta$ && \checkmark & & && & \checkmark \\
            FILO2-$\gamma$ && & \checkmark & && \checkmark & \\
            FILO2-$\delta$ && & \checkmark & && & \checkmark \\
            FILO2-$\mu$ && & & \checkmark  && \checkmark & \\
        \bottomrule
    \end{tabular}
    }
\end{table}

        For the experiment, as summarized in \Cref{table:variant-filo2-hybrid}, there are five hybrid variants of FILO2 in total. Variants $\alpha$, $\beta$, $\gamma$, and $\delta$ employ tabular binary classifiers, while variant $\mu$ leverages deep graph learning. The tabular classifier employs the Gradient Boosted Trees (GBT) and the Feedforward Neural Network (FNN) as baseline algorithms.  Given \Cref{eq:gbt} for GBT and \Cref{eq:fnn} for FNN, we set $\mathbb{M} = 100$ for GBT, and configure the FNN with $\mathbb{L} = 2$ hidden layers, each consisting of $256$ neurons. Threshold adjustments are applied to address class imbalance, which align with \Cref{eq:deterministic-th} and \Cref{eq:stochastic-th}. In FILO2, deterministic thresholds are set at $t = 0.8$ for both algorithms. Under stochastic thresholding, we use $t = 0.8$ with acceptance probabilities $p = 0.9$ for GBT and $p = 0.75$ for FNN. The graph model extend the graph \textit{Convolutional Network} (ConvNet) architecture \citep{joshi2019efficient,kool2022deep,dwivedi2023benchmarking} for direct binary solution's edge prediction. For graph model in FILO2, $t = 0.8$ is used for $\mathbb{X}$ instances and $t = 0.75$ for $\mathbb{B}$ instances. We also apply the tabu aspiration mechanism \citep{glover1997tabu} using a randomized threshold $p_\Theta$, allowing prohibited edges to performs move when $\mathcal{R} \sim \mathcal{U}(0,1)$ exceeds $p_\Theta$. For FILO2, we set $p_\Theta = 0.8$ for tabular models, and $p_\Theta = 0.6$ and $0.8$ for graph model on $\mathbb{X}$ and $\mathbb{B}$ respectively. 

    \paragraph{\textbf{Statistical analysis}}
        This analysis is conducted to assess the significant effect of the proposed mechanism. We performed the non-parametric one-tailed Wilcoxon signed-rank test \citep{JMLR:v7:demsar06a,arnold2021pils,accorsi2022guidelines,zarate2025machine} to compare the baseline and proposed hybrid mechanism, with Hypotheses \ref{hyp:a} and \ref{hyp:b} formulated as follows:
        \setcounter{hyp}{-1}
        \begin{hyp} \label{hyp:a}
            There is no statistically significant difference
        \end{hyp}
        \begin{hyp} \label{hyp:b}
            There is a statistically significant difference
        \end{hyp}

        In this research, we set $\alpha = 0.05$ \citep{arnold2021pils,zarate2025machine}. If the $p$-value $> \alpha$, we fail to reject \ref{hyp:a}, indicating that the average results of the two methods are not statistically different. If the $p$-value $\leq \alpha$, we reject \ref{hyp:a}, in which indicating that the average results of the proposed hybrid mechanism are statistically better compared with the baseline.

\begin{algorithm}[htbp]
    \footnotesize
    \caption{The Edge-Selector $f_{\theta}$ applied in FILO2.}
    \label{alg:filo2}
    \textbf{Input}: Problem instance: $P$\\
    \textbf{Output}: Solution: $S$
    \begin{algorithmic}[1] 
        \Procedure{Hybrid-FILO2}{$P$}
            \State $S_0 \leftarrow \Call{ConstructInitialSolution}{P}$
            \State $\mathcal{X} \gets \Call{FeatureExtraction}{S_0}$
            \State $S' \gets f_{\theta}(\mathcal{X})$   \Comment{\small edge-labeled solution}
            \State $k \leftarrow \Call{GreedyRouteEstimate}{P}$
            \If{$\text{Route}(S')>k$}
            \State $S' \leftarrow \Call{RouteMin}{S'}$
            \EndIf
            \State $S \leftarrow \Call{CoreOptimization}{S'}$
        \EndProcedure
    \end{algorithmic}
\end{algorithm}

    \subsection{Experiment with FILO2}
    \label{subsec:experiment-filo2}
        The pseudocode for the proposed hybrid edge selector mechanism integrated with the baseline FILO2 algorithm \citep{ACCORSI2024106562} is shown in \Cref{alg:filo2}. FILO2 begins with a construction phase that builds an initial feasible solution using a restricted version of the savings algorithm \citep{clarke1964scheduling,ARNOLD201932}. This is followed by an improvement phase, which may first attempt to reduce the number of routes if they are considered excessive. Subsequently, a core optimization process iteratively refines the solution through a localized optimization scheme. In our approach, the selector is applied after the initial solution $S_0$ is constructed. We embed $S_0$ and label each edge in the solution using the selector model $f_{\theta}$. These labels then guide the subsequent local search optimization phase as the algorithm refines $S_0$.

\begin{table}[htbp] 
    \begin{center}
        \caption{Summary of the average gap to the BKS over 5 runs on FILO2 (time in seconds).}
        \label{table:result-summary-filo2}
        \vspace*{0.2cm}
        \setlength\tabcolsep{4pt}
        \scalebox{0.78}{
            \begin{tabular}{l c cc c ccccccccccc}
                \toprule
                    \multirow{3}{*}{{\makecell{\textbf{Problem} \\ \textbf{Size}}}}
                    && \multicolumn{2}{c}{\textbf{Baseline}}
                    && \multicolumn{10}{c}{\textbf{Proposed Hybrid Mechanisms}}\\
                    \cmidrule(lr){3-4} \cmidrule(lr){5-15}
                    && \multicolumn{2}{c}{FILO2}
                    && \multicolumn{2}{c}{FILO2-$\alpha$}
                    &  \multicolumn{2}{c}{FILO2-$\beta$}
                    &  \multicolumn{2}{c}{FILO2-$\gamma$}
                    &  \multicolumn{2}{c}{FILO2-$\delta$}
                    &  \multicolumn{2}{c}{FILO2-$\mu$}\\
                    \cmidrule(lr){3-4} \cmidrule(lr){5-15}
                    && Gap (\%) & Time && Gap (\%) & Time & Gap (\%) & Time & Gap (\%) & Time & Gap (\%) & Time & Gap (\%) & Time\\
                \midrule
                    \multicolumn{15}{l}{\textbf{$\mathbb{X}$ Instances}} \\
                    101 - 200 && 0.138 & 91.26 && 0.138 & 68.36 & 0.138 & 69.66 & 0.131 & 68.00 & 0.143 & 67.32 & \textbf{0.114} & 117.73 \\
                    204 - 491 && \textbf{0.310} & 88.82 && 0.311 & 69.49 & \textbf{0.310} & 70.89 & 0.335 & 70.24 & 0.340 & 72.07 & 0.350 & 114.58 \\
                    502 - 1001 && 0.493 & 87.77 && 0.492 & 69.27 & 0.492 & 70.70 & 0.498 & 69.84 & 0.501 & 71.46 & \textbf{0.481} & 113.22 \\
                \midrule
                    Average Gap && \textbf{0.331} & 89.02 && \textbf{0.331} & 69.18 & \textbf{0.331} & 70.56 & 0.342 & 69.62 & 0.348 & 70.83 & 0.340 & 114.84 \\
                \midrule
                \midrule
                    \multicolumn{15}{l}{\textbf{$\mathbb{B}$ Instances}} \\
                    3k - 7k && 0.742 & 103.55 && 0.742 & 98.27 & 0.742 & 100.75 & 0.721 & 98.50 & 0.728 & 100.29 & \textbf{0.686} & 124.75 \\
                    10k - 16k && 1.262 & 109.15 && 1.262 & 106.02 & 1.262 & 110.05 & 1.274 & 105.63 & 1.294 & 108.50 & \textbf{1.241} & 144.00 \\
                    20k - 30k && 1.422 & 123.70 && 1.406 & 124.85 & 1.422 & 130.74 & 1.436 & 124.23 & 1.501 & 126.95 & \textbf{1.391} & 171.05 \\
                \midrule
                    Average Gap && 1.086 & 112.96 && 1.083 & 110.48 & 1.086 & 114.36 & 1.085 & 110.17 & 1.109 & 112.44 & \textbf{1.049} & 146.58 \\
                \bottomrule
            \end{tabular}
        }
    \end{center}
\end{table}

        \paragraph{\textbf{Computational results}}
            To assess the performance of the proposed hybrid mechanism compared with the baseline FILO2, we perform experiments utilizing both $\mathbb{X}$ and $\mathbb{B}$ instances to evaluate its effectiveness and scalability. \Cref{table:result-summary-filo2} presents a performance comparison between the baseline FILO2 and six variants of the proposed hybrid learning mechanisms. For the $\mathbb{X}$ instances, the hybrid variants generally maintain or slightly improve the gap compared to the baseline, while significantly reducing computational time, particularly those using a simple tabular architecture. Notably, FILO2-$\mu$, which is based on a GNN architecture, achieves the most significant performance improvement. In the $\mathbb{B}$ instances, the hybrid mechanisms again demonstrate competitive performance, with FILO2-$\mu$ consistently outperforming all other variants and the baseline across customer sizes. 
            
\begin{table}[htbp] 
    \begin{center}
        \caption{Average gap to BKS on $\mathbb{X}$ instances under different customer distribution.}
        \label{table:summary-filo2-customer-dist}
        \vspace*{0.2cm}
        \setlength\tabcolsep{6pt}
        \scalebox{0.82}{
            \begin{tabular}{l c c c c c c c}
                \toprule
                    \multirow{2}{*}{{\makecell{\textbf{Customer} \\ \textbf{Distribution}}}}
                    && \multicolumn{1}{c}{\textbf{Baseline}}
                    & \multicolumn{5}{c}{\textbf{Proposed Hybrid Mechanisms}} \\
                    \cmidrule(lr){3-3} \cmidrule(lr){4-8}
                    && FILO2 & FILO2-$\alpha$ & FILO2-$\beta$ & FILO2-$\gamma$ & FILO2-$\delta$ & FILO2-$\mu$ \\
                \midrule
                    Random (R) && 0.260 & \textbf{0.258} & \textbf{0.258} & 0.266 & 0.282 & 0.260 \\
                    Clustered (C) && 0.377 & 0.377 & 0.377 & 0.380 & 0.398 & \textbf{0.373} \\
                    Random-Clustered (RC) && \textbf{0.337} & \textbf{0.337} & \textbf{0.337} & 0.352 & 0.351 & 0.353 \\
                \bottomrule
            \end{tabular}
        }
    \end{center}
\end{table}
            
            However, this improvement comes with increased computation time, due to the overhead of model inference from the Deep Learning-based architecture. Among the tabular models, variants $\alpha$ and $\delta$ show marginal gains, but not to the extent observed with variant $\mu$. 

\begin{table}[htbp] 
    \begin{center}
        \caption{Detailed results variant $\mu$ in solving $\mathbb{B}$ instances (time in seconds).}
        \label{table:filo2-b-detailed-result}
        \vspace*{0.2cm}
        \setlength\tabcolsep{2.2pt}
        \scalebox{0.72}{
            \begin{tabular}{l rr rrr r rrr r}
                \toprule
                    \multirow{2}{*}{\textbf{Instance}} 
                    &&  \multicolumn{3}{c}{\textbf{FILO2}} 
                    &&  \multicolumn{3}{c}{\textbf{FILO2-$\mu$}} 
                    &&  \multirow{2}{*}{\textbf{BKS}} \\
                    \cmidrule(lr){3-5} \cmidrule(lr){7-9}
                    &&  \textbf{Avg (Gap)} & \textbf{Best (Gap)} & \textbf{Time} 
                    &&  \textbf{Avg (Gap)} & \textbf{Best (Gap)} & \textbf{Time} 
                    &&\\
                \midrule
                    Leuven1 && 193683 (0.433) & 193627 (0.404) & 90.4 && \textbf{193683 (0.414)} & 193576 (0.377) & 106.8 && 192848\\
                    Leuven2 && 112454.2 (0.954) & 112145 (0.677) & 121 && \textbf{112454.2 (0.767)} & 112027 (0.571) & 141.4 && 111391\\
                    Antwerp1 && 479808.4 (0.53) & 479551 (0.476) & 98.4 && \textbf{479808.4 (0.514)} & 479546 (0.475) & 119.8 && 477277\\
                    Antwerp2 && \textbf{294414.8 (1.052)} & 294173 (0.969) & 104.4 && \textbf{294414.8 (1.052)} & 294100 (0.944) & 131 && 291350\\
                    Ghent1 && \textbf{472596 (0.653)} & 472507 (0.634) & 105.4 && 472596 (0.669) & 472460 (0.624) & 133.2 && 469531\\
                    Ghent2 && 261645.2 (1.512) & 261346 (1.396) & 115.2 && \textbf{261645.2 (1.412)} & 261134 (1.314) & 149.4 && 257748\\
                    Brussels1 && 506761.8 (1.005) & 506609 (0.975) & 102.6 && \textbf{506761.8 (1.001)} & 506597 (0.972) & 142.4 && 501719\\
                    Brussels2 && \textbf{351955.2 (1.878)} & 351461 (1.735) & 113.4 && 351955.2 (1.882) & 351590 (1.772) & 151 && 345468\\
                    Flanders1 && 7294292 (0.748) & 7291040 (0.703) & 134.4 && \textbf{7294292 (0.708)} & 7289290 (0.679) & 184.6 && 7240118\\
                    Flanders2 && 4464854 (2.095) & 4462590 (2.043) & 144.4 && \textbf{4464854 (2.073)} & 4461230 (2.012) & 206.2 && 4373244\\
                \midrule
                    Average Gap && 1.086 &  & 112.96 && \textbf{1.049} & & 146.58 && \\
                    Median Gap && 0.980 & & && \textbf{0.884} & & && \\
                \bottomrule
            \end{tabular}
        }
    \end{center}
\end{table}

\begin{figure}[htbp]
    \centering
    \includegraphics[width=0.9\textwidth]{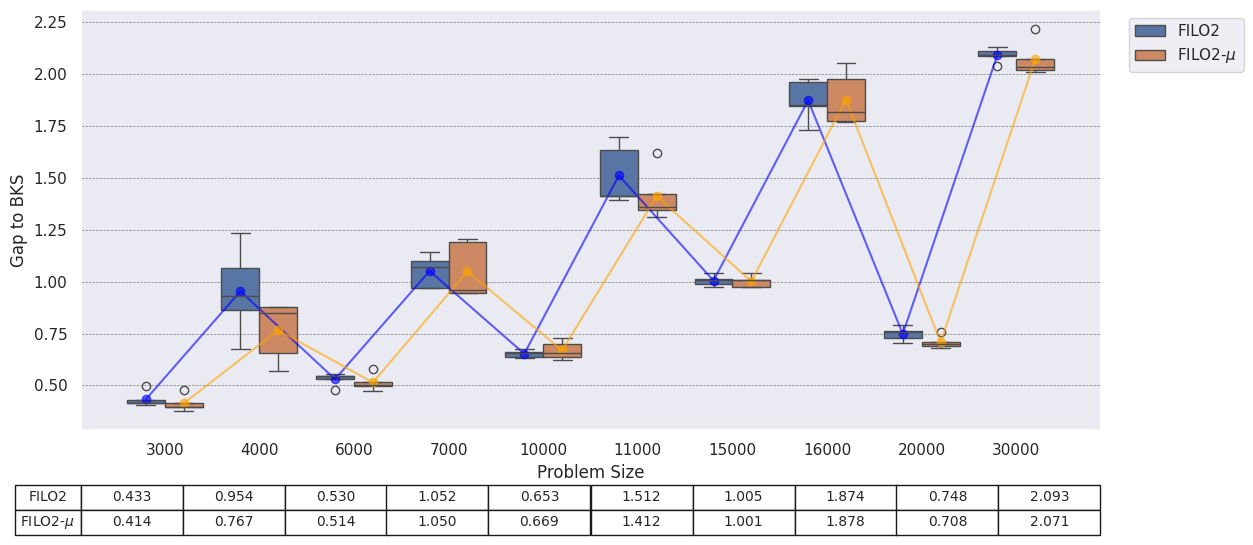}
    \caption{Comparison performance of FILO2 on $\mathbb{B}$ instances.} \label{fig:filo2-b}
\end{figure}

            Moreover, \Cref{table:summary-filo2-customer-dist} presents the average gap to the BKS on the $\mathbb{X}$ instances, categorized by different customer distribution: Random (R), Clustered (C), and Random-Clustered (RC). The goal is to evaluate how variations in customer spatial distribution affect the performance of the proposed hybrid learning-based mechanisms compared to the baseline FILO2 algorithm. In the Random (R) distribution, where customers are uniformly spread across the area, most hybrid variants (FILO2-$\alpha$, FILO2-$\beta$) slightly better than the baseline. FILO2-$\mu$ matches the baseline performance, while other variants show minor degradations. In the Clustered (C) case, the performance slightly worsens for most variants. However, FILO2-$\mu$ achieves the lowest gap of $0.373\%$, outperforming the baseline. Lastly, in the Random-Clustered (RC) scenario, FILO2-$\alpha$, and FILO2-$\beta$ all achieve matches the baseline performance. Furthermore, \Cref{table:filo2-b-detailed-result} presents a detailed comparison between the baseline FILO2 algorithm and the proposed hybrid variant FILO2-$\mu$ on all very large-scale instances from the $\mathbb{B}$ dataset. 

\begin{figure}[htbp]
    \centering
    \includegraphics[width=.8\textwidth]{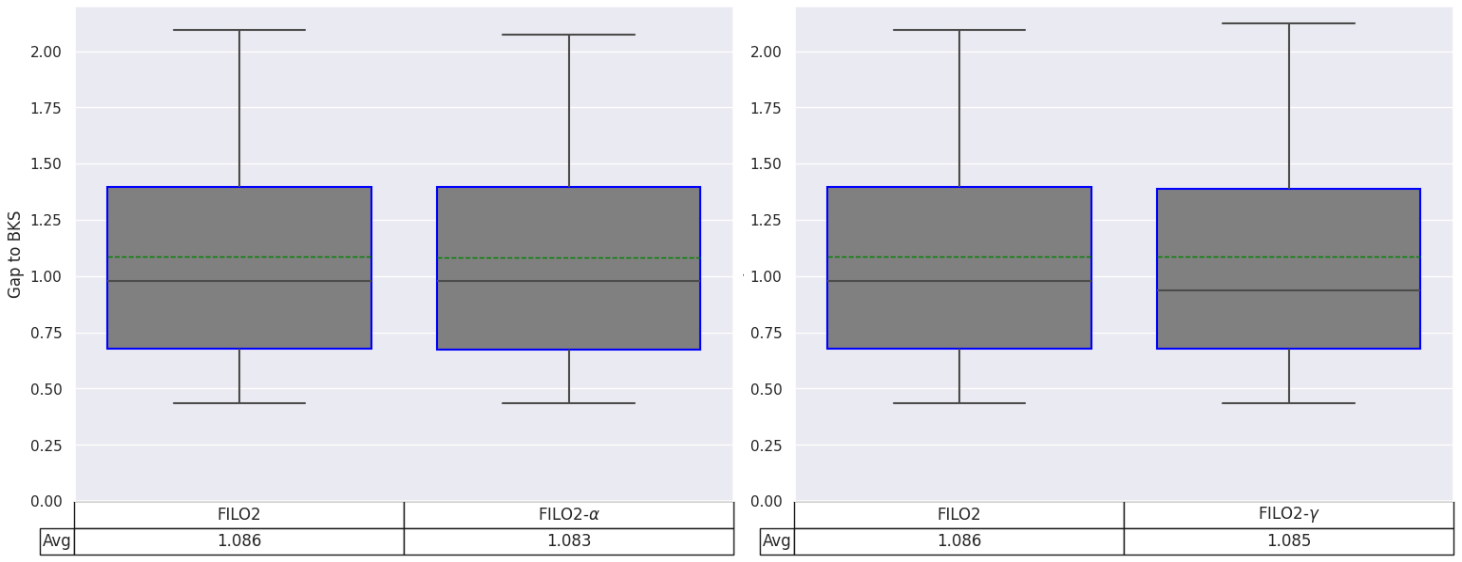}
    \caption{Boxplot of the performance of the hybrid FILO2 variants $\alpha$ and $\gamma$ on $\mathbb{B}$ instances.} \label{fig:filo2-b-tabular}
\end{figure}

\begin{figure}[htbp]
    \centering
    \includegraphics[width=.4\textwidth]{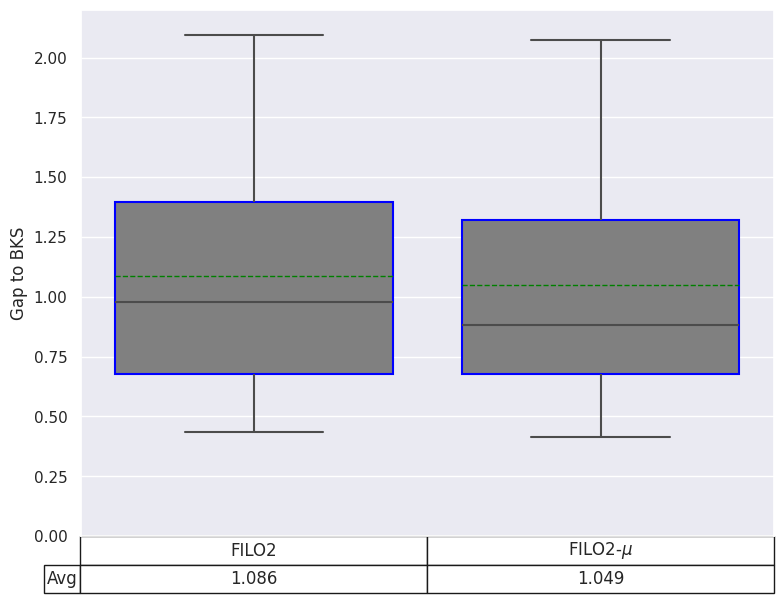}
    \caption{Boxplot of the performance of the hybrid FILO2 variants $\mu$ on $\mathbb{B}$ instances.} \label{fig:filo2-b-graph}
\end{figure}

            Overall, FILO2-$\mu$ consistently outperforms the baseline in terms of both average and best gap. These performance gains come with a slight increase in computational time, primarily due to the inference overhead introduced by the deep learning-based selector $f_{\theta}$ used in variant $\mu$.  Then, in \Cref{fig:filo2-b-tabular} for baseline FILO2, on $\mathbb{B}$ instances, the one-tailed Wilcoxon signed-rank test accept \Cref{hyp:a} with $p$-value $= 0.08986$ $(> 0.05)$ for variant $\alpha$ and $p$-value $= 0.5$ $(> 0.05)$ for variant $\gamma$, indicating that proposed hybrid variant $\alpha$ and $\gamma$ did not significantly improve the performance of the baseline FILO2 when solving $\mathbb{B}$ instances. Moreover, in \Cref{fig:filo2-b-graph} for baseline FILO2, on $\mathbb{B}$ instances, the one-tailed Wilcoxon signed-rank test rejects \Cref{hyp:a} with $p$-value $= 0.0186$ $(< 0.05)$, indicating that proposed hybrid variant $\mu$ significantly improve the baseline FILO2 when solving $\mathbb{B}$ instances.

\begin{algorithm}[t]
    \caption{The Edge-Selector $f_{\theta}$ variant $\mu$ applied in HGS.}
    \label{alg:hgs}
    \textbf{Input}: Problem instance: $P$\\
    \textbf{Output}: Solution: $S$
    \begin{algorithmic}[1]
        \Procedure{Hybrid-HGS}{$P$}
            \State $\mathbf{P} \gets \Call{InitialPopulation}{P}$ \Comment{\small random solutions improved by local search}
            \While{$t < T_{max}$}
                \State $S_0 \gets \Call{GetBestSol}{\mathbf{P}}$ \Comment{\small best feasible solution as the refereed solution}
                \State $\mathbf{P} \gets \mathbf{P} \cup f_{\theta}({S_0})$  \Comment{\small edge-labeled refereed solution in $\mathbf{P}$} 
                \State $(P_1,P_2) \gets \Call{SelectParentSolutions}{\mathbf{P}}$
                \State $C \gets \Call{Crossover}{P_1, P_2}$;
                \State $C' \gets \Call{LocalSearchImprovement}{C}$
                \State $\mathbf{P} \gets \mathbf{P} \cup C'$
                \If{$\neg\Call{Feasible}{C'}$}
                    \If{$\mathcal{R} < 0.5$} \Comment{\small random probability $50\%$}
                        \State $C'' \gets \Call{LocalSearch}{C'}$
                        \State $\mathbf{P} \gets \mathbf{P} \cup C''$
                    \EndIf
                \EndIf
                
                \If{$\Call{Size}{\mathbf{P}} > \mu$} \Comment{\small maximum subpopulation size reached}
                    \State $\Call{PruneSurvivorsBasedOnFitness}{\mathbf{P}}$
                    \State $\Call{PenaltyAdjustment}{\mathbf{P}}$
                \EndIf
                \State $\Call{PenaltyAdjustment}{\mathbf{P}}$
            \EndWhile
        \EndProcedure
    \end{algorithmic}
\end{algorithm}
            
    \subsection{Generalization Baseline HGS}
    \label{subsec:experiment-hgs} 
        To assess the generalization capability of the proposed hybrid mechanism, we applied the $\mu$ variant in HGS to solve the $\mathbb{X}$ instances. This is because, as shown in \Cref{subsec:experiment-filo2}, the $\mu$ variant is the only one that achieves a statistically significant improvement over FILO2, particularly when solving very large instances. Therefore, we explored the generalizability of the $\mu$ variant using a different baseline algorithm. As described in \Cref{sec:baseline-metaheuristics}, FILO2 and HGS are structurally very different. HGS \citep{vidal2022hybrid} evolves a diverse population of solutions that comprises both feasible and infeasible individuals, such that $\mathbf{P} = \mathbf{P}_{\text{feas}} \cup \mathbf{P}_{\text{infeas}}$, and improves solution quality via recombination and local search operations. 
        
        In contrast, FILO2 refines a single incumbent solution through successive local search steps, using a simulated annealing acceptance criterion to maintain search diversity \citep{accorsi2021fast}. These structural differences provide a basis for testing the generalization ability of the proposed mechanism across different metaheuristic baselines. The pseudocode for the proposed hybrid edge selector mechanism integrated with the baseline HGS algorithm is presented in \Cref{alg:hgs}. After an initial population is generated, the algorithm iteratively produces new solutions by selecting two parents, $P_1$ and $P_2$, recombining them to create a new solution $C$, improving this solution via local search, and inserting the result back into the population. 
        
        This process is repeated until a termination criterion is met, which, in this research, is defined by a predefined time limit $T_{\text{max}}$ \citep{vidal2022hybrid}. In this framework, the selector marks the best feasible solution from the initial population as the reference solution $S_0$. This solution is then input into the selector $f_{\theta}$, where all edges in $S_0$ are labeled to guide the local search for improving all solutions in $\mathbf{P}$. Furthermore, whenever a specified number of consecutive iterations $N_{\text{IT}}$ without improvement is reached, the algorithm regenerates the population. At this point, the selector $f_{\theta}$ also performs predictions again using a new reference solution from the newly generated population. Prior to conducting experiments on the $\mathbb{X}$ instances, we set the threshold parameters as follows: $t = 0.75$ and $p_\Theta = 0.7$ for instances with $N < 500$, and $t = 0.85$ for larger instances. The experimental setup follows the configuration used in HGS \citep{vidal2022hybrid}, with the maximum runtime defined as $T_{\text{max}} = N \times 2.4$ seconds for all $\mathbb{X}$ instances.

\begin{table}[htbp]
    \begin{center}
        \caption{Summary performances variant $\mu$ on HGS with $T_{max} = N \times \nicefrac{240}{100}$.}
        \label{table:generalized-hgs-summary}
        \vspace*{0.2cm}
        \setlength\tabcolsep{4pt}
        \scalebox{0.8}{
            \begin{tabular}{ll cc c cc}
                \toprule
                    \multirow{2}{*}{\makecell{\textbf{Problem} \\ 
                    \textbf{Size}}} 
                    && \multicolumn{2}{c}{\textbf{HGS}} 
                    && \multicolumn{2}{c}{\textbf{HGS-$\mu$}} \\
                    \cmidrule(lr){3-4} \cmidrule(lr){6-7}
                    && \textbf{Avg Gap} & \textbf{Best Gap} 
                    && \textbf{Avg Gap} & \textbf{Best Gap} \\
                \midrule
                    101 - 200 && \textbf{0.005} & 0.002 && \textbf{0.005} & 0.001 \\
                    204 - 491 && 0.071 & 0.039 && \textbf{0.061} & 0.032 \\
                    502 - 1001 && 0.268 & 0.199 && \textbf{0.239} & 0.173 \\
                \midrule
                    Average Gap && 0.120 & && \textbf{0.105} & \\
                    Median Gap && 0.059 & && \textbf{0.046} & \\
                \bottomrule
            \end{tabular}
        }
    \end{center}
\end{table}

\begin{figure}[htbp]
    \centering
    \includegraphics[width=1\textwidth]{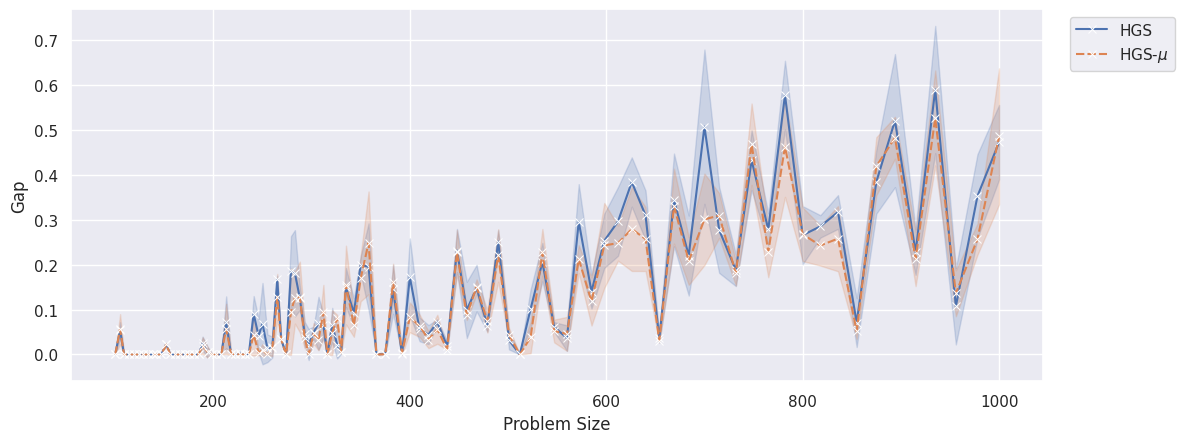}
    \caption{Comparison performance of HGS on $\mathbb{X}$ instances.} \label{fig:hgs-x} 
\end{figure}

        \paragraph{\textbf{Computational results}}
            \Cref{table:generalized-hgs-summary} presents a comparative performance summary between the baseline HGS and the proposed hybrid variant HGS-$\mu$ that incorporates the graph learning as the baseline learning model (as summarized in \Cref{table:variant-filo2-hybrid}, similar with baseline FILO2). The table reports results over three instance size categories from the $\mathbb{X}$ benchmark set, where each configuration was evaluated under a similar maximum computational time $T_{\text{max}} = N \times \frac{240}{100}$. 

\begin{table}[htbp]
    \begin{center}
        \vspace*{-0.2cm}
        \caption{Performance on $\mathbb{X}$ instances with varying customer distributions.}
        \vspace*{0.3cm}
        \setlength\tabcolsep{6pt}
        \label{table:result-hgs-customer-dist}
        \scalebox{0.8}{
            \begin{tabular}{l r r}
                \toprule
                \textbf{Customer Distribution} & \textbf{HGS} & \textbf{HGS-$\mu$} \\
                \midrule
                Random (R) & 0.105 & \textbf{0.092} \\
                Clustered (C) & 0.110 & \textbf{0.098 }\\
                Random-Clustered (RC) & 0.128 & \textbf{0.113} \\
                \bottomrule
            \end{tabular}
        }
    \end{center}
\end{table}

\begin{figure}[htbp]
    \centering
    \includegraphics[width=.52\textwidth]{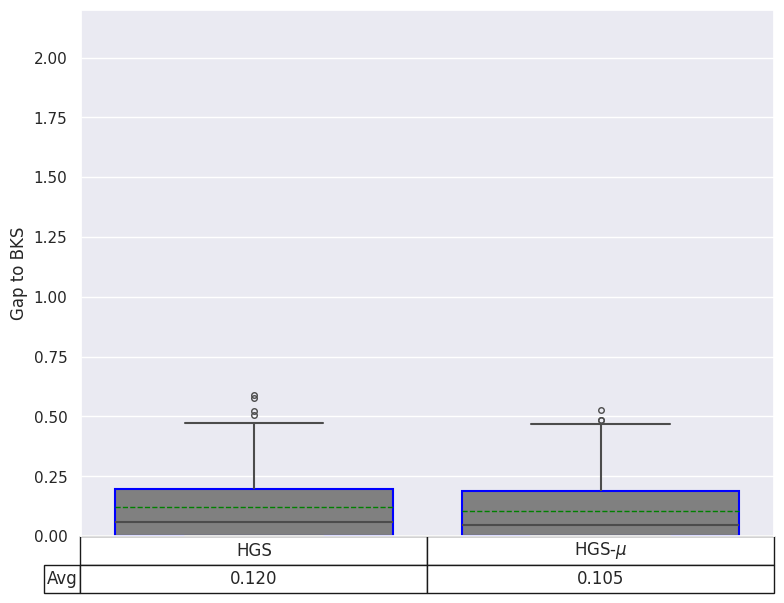}
    \caption{Boxplot of the performance of the hybrid HGS variants $\mu$ on $\mathbb{X}$ instances.} \label{fig:hgs-x-graph}
\end{figure}

            Across all problem sizes, HGS-$\mu$ almost consistently improves the solution quality compared to the baseline HGS. For medium-sized instances (with $204$ to $491$ nodes), the average gap is reduced from $0.071\%$ to $0.061\%$, and for large instances (with $502$ to $1001$ nodes), the improvement is more pronounced, with the average gap reduced from $0.268\%$ to $0.239\%$. While both methods perform nearly similar on small instances, HGS-$\mu$ achieves a slightly better in best resulted solution. The overall average gap across all instance sizes is improved, indicating an enhancement in solution quality. Further, \Cref{table:result-hgs-customer-dist} presents a comparative evaluation of the baseline HGS and the proposed hybrid variant HGS-$\mu$, across different customer distribution patterns on $\mathbb{X}$ benchmark instances. Across all cases, HGS-$\mu$ improves the baseline HGS. Specifically, in Random instances, HGS-$\mu$ achieves a gap, improving upon the baseline. For Clustered distributions, HGS-$\mu$ also yields a better gap, compared the baseline. As shown in \Cref{fig:hgs-x-graph}, for the baseline HGS on the $\mathbb{X}$ instances, the one-tailed Wilcoxon signed-rank test rejects \Cref{hyp:a} with a $p$-value of $0.000056$ $(< 0.05)$, suggesting that the proposed hybrid variant~$\mu$ statistically improves the performance of the baseline HGS on these instances. Furthermore, we compare the proposed method with other state-of-the-art metaheuristic solvers for the CVRP.

            As summarized in \Cref{table:result-stoa-comparison}, we evaluate its performance against several well-known solvers: LKH\footnote{LKH-3.0.9: \url{http://webhotel4.ruc.dk/~keld/research/LKH-3/}} \citep{helsgaun2017extension}, SISRs\footnote{Results reported in \cite{vidal2022hybrid}} \citep{christiaens2020slack}, HGS\footnote{HGS-CVRP: \url{https://github.com/vidalt/HGS-CVRP}} \citep{vidal2022hybrid}, and the general commercial solver Hexaly\footnote{Hexaly 13.0: \url{https://www.hexaly.com}}. All algorithms were evaluated under equivalent stopping criteria, with the maximum runtime set to $T_{\text{max}} = N \times \nicefrac{240}{100}$ seconds.

\begin{table}[htbp] 
    \begin{center}
        \vspace*{-0.2cm}
        \caption{Comparison of solution quality with $T_{\text{max}} = N \times \nicefrac{240}{100}$ on $\mathbb{X}$ instances.}
        \vspace*{0.3cm}
        \setlength\tabcolsep{6pt}
        \label{table:result-stoa-comparison}
        \scalebox{0.92}{
            \begin{tabular}{l r r r r r}
                \toprule
                \textbf{Measurement} & \textbf{LKH-3} & \textbf{SISRs} & \textbf{Hexaly} & \textbf{HGS} & \textbf{HGS-$\mu$}\\
                \midrule
                Average Gap & 1.029 & 0.194 & 1.731 & \textit{0.120} & \textbf{0.105}\\
                Median Gap & 0.898 & 0.170 & 1.120 & \textit{0.059} & \textbf{0.049}\\
                \bottomrule
            \end{tabular}
        }
    \end{center}
\end{table}

        From \Cref{table:result-stoa-comparison}, it can be observed that although HGS already outperforms other metaheuristic solvers, the proposed hybrid mechanism achieves even better performance than HGS and indeed surpasses all other compared solvers. This demonstrates the capability of the proposed hybrid mechanism further to enhance the performance of a state-of-the-art baseline metaheuristic.

    \subsection{Generalization for VRP with Time Windows}
    \label{subsec:experiment-hgs-tw}
        To further evaluate the generalizability of the proposed hybrid mechanism, we conducted experiments on a more complex variant of the \textit{Vehicle Routing Problems} (VRPs), namely the Capacitated VRP with Time Windows (CVRPTW) \citep{solomon1987algorithms}. For this purpose, we integrated the proposed selector $f_{\theta}$ into the baseline HGS-TW \citep{kool2022hybrid}. Since the baseline HGS-TW already demonstrates strong performance on small- and medium-sized instances \citep{kool2022hybrid}, our focus is placed on testing the hybrid variant $\mu$ on a subset of the Gehring \& Homberger benchmark instances \citep{gehring1999parallel}. Specifically, we selected large-scale instances containing $800$ and $1,000$ customer nodes from the C, R, and RC sets, which are categorized based on customer distribution. In this setup, we applied variant $\mu$ within HGS-TW to solve the selected CVRPTW instances, using a decision threshold of $t = 0.85$ and an aspiration parameter of $p_\Theta = 0.6$. The integration mechanism mirrors the one used for the CVRP experiments: the best feasible solution from the initial population is selected as the reference solution $S_0$, which is then passed to the selector $f_{\theta}$. 
        Additionally, if a predefined number of non-improving iterations $N_{\text{IT}}$ is reached, the population is regenerated, and the selector $f_{\theta}$ re-evaluates a new reference solution from this updated population.

\begin{table}[htbp]
    \begin{center}
        \caption{Summary performances variant $\mu$ on HGS-TW with $T_{\text{max}} = N \times \nicefrac{240}{100}$.}
        \label{table:generalized-hgs-tw-summary}
        \vspace*{0.2cm}
        \setlength\tabcolsep{4pt}
        \scalebox{0.86}{
            \begin{tabular}{ll cc c cc}
                \toprule
                    \multirow{2}{*}{\makecell{\textbf{Problem} \\ 
                    \textbf{Size}}} 
                    && \multicolumn{2}{c}{\textbf{HGS-TW}} 
                    && \multicolumn{2}{c}{\textbf{HGS-TW-$\mu$}} \\
                    \cmidrule(lr){3-4} \cmidrule(lr){6-7}
                    && \textbf{Avg Gap} & \textbf{Best Gap} 
                    && \textbf{Avg Gap} & \textbf{Best Gap} \\
                \midrule
                    800 && 0.509 & 0.353 && \textbf{0.437} & 0.304 \\
                    1000 && 0.790 & 0.578 && \textbf{0.675} & 0.497 \\
                \midrule
                    Average Gap && 0.649 & && \textbf{0.556} & \\
                    Median Gap && 0.628 & && \textbf{0.547} & \\
                \bottomrule
            \end{tabular}
        }
    \end{center}
\end{table}

        \paragraph{\textbf{Computational results}}
            The experimental results summarized in \Cref{table:generalized-hgs-tw-summary} and \Cref{table:result-hgs-tw-customer-dist} highlight the effectiveness of the hybrid $\mu$ variant when integrated into the baseline HGS-TW for solving Gehring \& Homberger instances \citep{gehring1999parallel}. As shown in \Cref{table:generalized-hgs-tw-summary}, HGS-TW-$\mu$ consistently outperforms the baseline across large problem sizes. Additionally, \Cref{table:result-hgs-tw-customer-dist} demonstrates the method's robustness across different customer distributions. The hybrid approach yields improvements in all scenarios, with particularly notable reductions in the average gap for Random (R) and Random-Clustered (RC) distributions. 

\begin{table}[htbp]
    \begin{center}
        \vspace*{-0.2cm}
        \caption{Performance of HGS-TW with varying customer distributions.}
        \vspace*{0.3cm}
        \setlength\tabcolsep{6pt}
        \label{table:result-hgs-tw-customer-dist}
        \scalebox{0.82}{
            \begin{tabular}{l r r}
                \toprule
                \textbf{Customer Distribution} & \textbf{HGS-TW} & \textbf{HGS-TW-$\mu$} \\
                \midrule
                Random (R) & 0.923 & \textbf{0.765} \\
                Clustered (C) & 0.184 & \textbf{0.153}\\
                Random-Clustered (RC) & 0.841 & \textbf{0.749} \\
                \bottomrule
            \end{tabular}
        }
    \end{center}
\end{table}

\begin{figure}[htbp]
    \centering
    \includegraphics[width=1\textwidth]{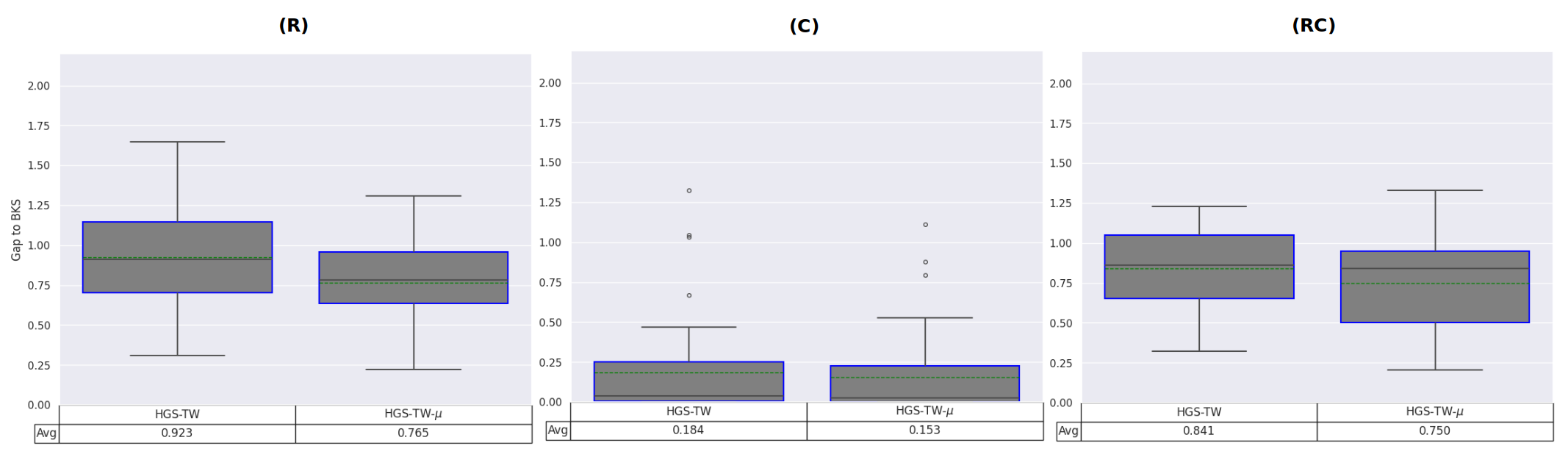}
    \caption{Comparison performance of HGS-TW.} \label{fig:hgs-tw}
\end{figure}

        As shown in \Cref{table:generalized-hgs-tw-summary}, HGS-TW-$\mu$ outperforms the baseline across large problem sizes. Additionally, \Cref{table:result-hgs-tw-customer-dist} demonstrates the improvement across different customer distributions. The hybrid approach yields improvements in all scenarios, with particularly notable reductions in the average gap for Random (R) and Random-Clustered (RC) distributions. As shown in \Cref{table:result-hgs-tw-customer-dist}, the one-tailed Wilcoxon signed-rank test rejects \Cref{hyp:a} for the baseline HGS-TW on a subset of the Gehring \& Homberger instances \citep{gehring1999parallel}. Specifically, statistically significant improvements were observed with $p$-values of $0.000002$ ($< 0.05$) for the random distribution, $0.00177$ ($< 0.05$) for the clustered distribution, and $0.00006$ ($< 0.05$) for the random-clustered distribution. These results demonstrate that the proposed hybrid variant $\mu$ significantly enhances the performance of the baseline across all types of customer distributions.

\newpage
\section{Conclusion} 
\label{sec:conclusion} 
    In this paper, we introduced a novel iterative learning-based hybrid optimization framework aimed at enhancing local search performance within metaheuristic solvers. At the core of the proposed approach is an edge selector model that classifies solution edges to identify and prohibit unpromising moves, thereby reducing the neighborhood search space and filtering out unnecessary local search operations. This model operates within an in-loop, machine learning-guided optimization framework that enables adaptive, data-driven guidance through repeated decision inference. We implemented the selector using two learning paradigms: tabular binary classifiers (GBT and FNN) and Graph Neural Network (GNN), with both deterministic and stochastic thresholding techniques applied to address class imbalance. To evaluate scalability, we applied the framework to CVRP instances with customer sizes ranging from 100 to 30,000. The results show statistically significant improvements for both medium and large scale problem instances. To assess generalizability, we integrated the mechanism into two structurally distinct metaheuristics, HGS as a population-based method and FILO2 as a single-solution-based method. Both showed consistent and statistically significant performance gains when augmented with the hybrid mechanism. Finally, we extended the approach to solve CVRPTW using HGS-TW as the baseline. The results demonstrated that the hybrid mechanism significantly improves the solver’s performance across all customer distribution types on large instances, confirming the robustness of the proposed approach.

\newpage
\appendix

\section{Appendix}

\subsection{Edge-Solution Features for Tabular Learning}
\label{sec:app-feature}
    Let $R$ denote the set of all routes in a given solution. Each route, denoted as $r_k = \{ c_1, c_2, \dots, c_k \} \in R$, represents an ordered sequence of customer nodes that must be visited consecutively. The relationship between any two nodes $c_i$ and $c_j$ in terms of their proximity within the problem space is captured by the neighborhood rank $\mathcal{R}_{ij}$, which reflects the relative closeness of node $c_j$ to node $c_i$ based on distance or similarity metrics. Each route $r_k \in R$ consists of a series of directed edges forming the path $(D, c_1), (c_1, c_2), \dots, (c_k, D) \in E$, where $D$ denotes the depot serving as both the starting and ending point of the route. The Euclidean distance between any two customer nodes $c_j$ and $c_k$ is denoted by $d(c_j, c_k)$, computed using their spatial coordinates. Specifically, the horizontal and vertical positions of node $c_k$ are denoted by $x(c_k)$ and $y(c_k)$, respectively. In addition, each customer node $c_j$ is associated with a demand value, represented as $q(c_j)$, which indicates the quantity of goods required by that customer. If nodes $c_i$ and $c_j$ are connected by an edge in a route $r_k \in R$ within a solution $S$, then this edge can be characterized by a set of solution-specific features, including:
    \begin{itemize}
        \item percentage load, $x_1$
        \begin{equation}
            x_1 := \dfrac{\left(q(c_i) + q(c_j) \right)}{\sum_{N} q(c)}
        \end{equation}
        \item utilization edge, $x_2$
        \begin{equation}
            x_2 := \dfrac{\left(q(c_i) + q(c_j) \right)}{\sum_{R_k} q(c)}
        \end{equation}
        \item rank of the neighborhood, $x_3$
        \begin{equation}
            x_3 := \mathcal{R}_{ij} \cdot \delta_{\mathcal{R}_{ij} \leq \Gamma} + (-1) \cdot (1-\delta_{\mathcal{R}_{ij} \leq \Gamma})
        \end{equation}
        \item percentage distance, $x_4$
        \begin{equation}
            x_4 := \dfrac{d\left(c_i, c_j \right)}{S}
        \end{equation}
    \end{itemize}
    The term $\delta_{\mathcal{R}_{ij} \leq \Gamma}$ is used for extending the Kronecker delta notation to include a special condition, $\delta_{\text{cond}}$, which equal to $1$ if the condition is met and $0$ otherwise. For $x_3$, rank of the neighborhood, we adjust the granularity $\Gamma$ of the baseline algorithm. To ensure this, the value of $x_3$ will be set to $-1$ whenever $\mathcal{R}_{ij} > 25$. This transformation respects the granularity mechanism of the local search used in the baseline algorithm while maintain the consistency in feature representation.

\newpage

\subsection{Detailed Experiment using Baseline HGS}
\label{sec:app-detailed-hgs}

\begin{table}[htbp]
    \begin{center}
        \caption{Detailed performances variant $\mu$ with HGS on $\mathbb{X}$ instances.}
        \label{table:generalized-hgs-detailed-1}
        \vspace*{0.2cm}
        \setlength\tabcolsep{4pt}
        \scalebox{0.68}{
            \begin{tabular}{ll rr r rr rr}
                \toprule
                    \multirow{2}{*}{\textbf{Instance}} 
                    && \multicolumn{2}{c}{\textbf{HGS}} 
                    && \multicolumn{2}{c}{\textbf{HGS-$\mu$}}
                    && \multirow{2}{*}{\textbf{BKS}}\\
                    \cmidrule(lr){3-4} \cmidrule(lr){6-7}
                    && \textbf{Avg Gap} & \textbf{Best Gap} 
                    && \textbf{Avg Gap} & \textbf{Best Gap} \\
                \midrule
                    X-n101-k25 && 27591 (0.0000) & 27591 (0.0000) && 27591 (0.0000) & 27591 (0.0000) && 27591 \\
                    X-n106-k14 && 26376.2 (0.0539) & 26370 (0.0303) && 26376.8 (0.0561) & 26362 (0.0000) && 26362 \\
                    X-n110-k13 && 14971 (0.0000) & 14971 (0.0000) && 14971 (0.0000) & 14971 (0.0000) && 14971 \\
                    X-n115-k10 && 12747 (0.0000) & 12747 (0.0000) && 12747 (0.0000) & 12747 (0.0000) && 12747 \\
                    X-n120-k6 && 13332 (0.0000) & 13332 (0.0000) && 13332 (0.0000) & 13332 (0.0000) && 13332 \\
                    X-n125-k30 && 55539 (0.0000) & 55539 (0.0000) && 55539 (0.0000) & 55539 (0.0000) && 55539 \\
                    X-n129-k18 && 28940 (0.0000) & 28940 (0.0000) && 28940 (0.0000) & 28940 (0.0000) && 28940 \\
                    X-n134-k13 && 10916 (0.0000) & 10916 (0.0000) && 10916 (0.0000) & 10916 (0.0000) && 10916 \\
                    X-n139-k10 && 13590 (0.0000) & 13590 (0.0000) && 13590 (0.0000) & 13590 (0.0000) && 13590 \\
                    X-n143-k7 && 15700 (0.0000) & 15700 (0.0000) && 15700 (0.0000) & 15700 (0.0000) && 15700 \\
                    X-n148-k46 && 43448 (0.0000) & 43448 (0.0000) && 43448 (0.0000) & 43448 (0.0000) && 43448 \\
                    X-n153-k22 && 21225 (0.0236) & 21225 (0.0236) && 21224.8 (0.0226) & 21224 (0.0189) && 21220 \\
                    X-n157-k13 && 16876 (0.0000) & 16876 (0.0000) && 16876 (0.0000) & 16876 (0.0000) && 16876 \\
                    X-n162-k11 && 14138 (0.0000) & 14138 (0.0000) && 14138 (0.0000) & 14138 (0.0000) && 14138 \\
                    X-n167-k10 && 20557 (0.0000) & 20557 (0.0000) && 20557 (0.0000) & 20557 (0.0000) && 20557 \\
                    X-n172-k51 && 45607 (0.0000) & 45607 (0.0000) && 45607 (0.0000) & 45607 (0.0000) && 45607 \\
                    X-n176-k26 && 47812 (0.0000) & 47812 (0.0000) && 47812 (0.0000) & 47812 (0.0000) && 47812 \\
                    X-n181-k23 && 25569 (0.0000) & 25569 (0.0000) && 25569 (0.0000) & 25569 (0.0000) && 25569 \\
                    X-n186-k15 && 24145 (0.0000) & 24145 (0.0000) && 24145 (0.0000) & 24145 (0.0000) && 24145 \\
                    X-n190-k8 && 16984.2 (0.0247) & 16980 (0.0000) && 16983.4 (0.0200) & 16980 (0.0000) && 16980 \\
                    X-n195-k51 && 44227.4 (0.0054) & 44225 (0.0000) && 44227.4 (0.0054) & 44225 (0.0000) && 44225 \\
                    X-n200-k36 && 58578 (0.0000) & 58578 (0.0000) && 58578 (0.0000) & 58578 (0.0000) && 58578 \\
                    X-n204-k19 && 19565 (0.0000) & 19565 (0.0000) && 19565 (0.0000) & 19565 (0.0000) && 19565 \\
                    X-n209-k16 && 30656 (0.0000) & 30656 (0.0000) && 30656 (0.0000) & 30656 (0.0000) && 30656 \\
                    X-n214-k11 && 10863.8 (0.0718) & 10856 (0.0000) && 10862.2 (0.0571) & 10859 (0.0276) && 10856 \\
                    X-n219-k73 && 117595 (0.0000) & 117595 (0.0000) && 117597.2 (0.0019) & 117595 (0.0000) && 117595 \\
                    X-n223-k34 && 40437 (0.0000) & 40437 (0.0000) && 40437 (0.0000) & 40437 (0.0000) && 40437 \\
                    X-n228-k23 && 25743 (0.0039) & 25743 (0.0039) && 25742.2 (0.0008) & 25742 (0.0000) && 25742 \\
                    X-n233-k16 && 19230 (0.0000) & 19230 (0.0000) && 19230 (0.0000) & 19230 (0.0000) && 19230 \\
                    X-n237-k14 && 27042 (0.0000) & 27042 (0.0000) && 27042 (0.0000) & 27042 (0.0000) && 27042 \\
                    X-n242-k48 && 82825.6 (0.0901) & 82768 (0.0205) && 82787.6 (0.0442) & 82768 (0.0205) && 82751 \\
                    X-n247-k50 && 37289.6 (0.0419) & 37278 (0.0107) && 37275.4 (0.0038) & 37274 (0.0000) && 37274 \\
                    X-n251-k28 && 38710.6 (0.0688) & 38684 (0.0000) && 38687.8 (0.0098) & 38684 (0.0000) && 38684 \\
                    X-n256-k16 && 18841.6 (0.0138) & 18839 (0.0000) && 18839 (0.0000) & 18839 (0.0000) && 18839 \\
                    X-n261-k13 && 26562.4 (0.0166) & 26558 (0.0000) && 26561.6 (0.0136) & 26558 (0.0000) && 26558 \\
                    X-n266-k58 && 75604.2 (0.1672) & 75595 (0.1550) && 75575 (0.1285) & 75517 (0.0517) && 75478 \\
                    X-n270-k35 && 35303 (0.0340) & 35303 (0.0340) && 35303 (0.0340) & 35303 (0.0340) && 35291 \\
                    X-n275-k28 && 21245 (0.0000) & 21245 (0.0000) && 21245 (0.0000) & 21245 (0.0000) && 21245 \\
                    X-n280-k17 && 33565.2 (0.1857) & 33536 (0.0985) && 33535 (0.0955) & 33506 (0.0090) && 33503 \\
                    X-n284-k15 && 20251.6 (0.1811) & 20217 (0.0099) && 20240.6 (0.1266) & 20228 (0.0643) && 20215 \\
                    X-n289-k60 && 95270.8 (0.1259) & 95238 (0.0914) && 95273 (0.1282) & 95183 (0.0336) && 95151 \\
                    X-n294-k50 && 47182.2 (0.0450) & 47167 (0.0127) && 47178.4 (0.0369) & 47161 (0.0000) && 47161 \\
                    X-n298-k31 && 34238.6 (0.0222) & 34231 (0.0000) && 34231 (0.0000) & 34231 (0.0000) && 34231 \\
                    X-n303-k21 && 21746.6 (0.0488) & 21743 (0.0322) && 21746.2 (0.0469) & 21739 (0.0138) && 21736 \\
                \bottomrule   
            \end{tabular}
        }
    \end{center}
\end{table}

\newpage

\begin{table}[htbp]
    \begin{center}
        \caption{Detailed performances variant $\mu$ with HGS on $\mathbb{X}$ instances (continue).}
        \label{table:generalized-hgs-detailed-2}
        \vspace*{0.2cm}
        \setlength\tabcolsep{4pt}
        \scalebox{0.68}{
            \begin{tabular}{ll rr r rr rr}
                \toprule
                    \multirow{2}{*}{\textbf{Instance}} 
                    && \multicolumn{2}{c}{\textbf{HGS}} 
                    && \multicolumn{2}{c}{\textbf{HGS-$\mu$}}
                    && \multirow{2}{*}{\textbf{BKS}}\\
                    \cmidrule(lr){3-4} \cmidrule(lr){6-7}
                    && \textbf{Avg Gap} & \textbf{Best Gap} 
                    && \textbf{Avg Gap} & \textbf{Best Gap} \\
                \midrule
                X-n308-k13 && 25877 (0.0696) & 25862 (0.0116) && 25867.2 (0.0317) & 25862 (0.0116) && 25859 \\
                X-n313-k71 && 94104.2 (0.0651) & 94085 (0.0447) && 94132.2 (0.0949) & 94044 (0.0011) && 94043 \\
                X-n317-k53 && 78355 (0.0000) & 78355 (0.0000) && 78355 (0.0000) & 78355 (0.0000) && 78355 \\
                X-n322-k28 && 29853.6 (0.0657) & 29834 (0.0000) && 29848.6 (0.0489) & 29834 (0.0000) && 29834 \\
                X-n327-k20 && 27538 (0.0218) & 27532 (0.0000) && 27554.4 (0.0814) & 27553 (0.0763) && 27532 \\
                X-n331-k15 && 31103.4 (0.0045) & 31103 (0.0032) && 31103.4 (0.0045) & 31103 (0.0032) && 31102 \\
                X-n336-k84 && 139318.8 (0.1494) & 139249 (0.0992) && 139326.4 (0.1548) & 139192 (0.0582) && 139111 \\
                X-n344-k43 && 42089.2 (0.0932) & 42072 (0.0523) && 42077.4 (0.0652) & 42061 (0.0262) && 42050 \\
                X-n351-k40 && 25947.6 (0.1993) & 25939 (0.1660) && 25940 (0.1699) & 25922 (0.1004) && 25896 \\
                X-n359-k29 && 51605.8 (0.1957) & 51571 (0.1281) && 51633.2 (0.2489) & 51578 (0.1417) && 51505 \\
                X-n367-k17 && 22814 (0.0000) & 22814 (0.0000) && 22814 (0.0000) & 22814 (0.0000) && 22814 \\
                X-n376-k94 && 147716 (0.0020) & 147713 (0.0000) && 147714 (0.0007) & 147713 (0.0000) && 147713 \\
                X-n384-k52 && 66035.8 (0.1453) & 65975 (0.0531) && 66046.6 (0.1617) & 66003 (0.0955) && 65940 \\
                X-n393-k38 && 38260 (0.0000) & 38260 (0.0000) && 38261.2 (0.0031) & 38260 (0.0000) && 38260 \\
                X-n401-k29 && 66277.4 (0.1729) & 66213 (0.0756) && 66217.8 (0.0828) & 66180 (0.0257) && 66163 \\
                X-n411-k19 && 19723 (0.0558) & 19719 (0.0355) && 19724.6 (0.0639) & 19718 (0.0304) && 19712 \\
                X-n420-k130 && 107845.8 (0.0443) & 107818 (0.0186) && 107833.6 (0.0330) & 107824 (0.0241) && 107798 \\
                X-n429-k61 && 65496.4 (0.0724) & 65490 (0.0626) && 65485.8 (0.0562) & 65464 (0.0229) && 65449 \\
                X-n439-k37 && 36397.8 (0.0187) & 36395 (0.0110) && 36395 (0.0110) & 36395 (0.0110) && 36391 \\
                X-n449-k29 && 55361 (0.2317) & 55343 (0.1992) && 55359.2 (0.2285) & 55334 (0.1829) && 55233 \\
                X-n459-k26 && 24163 (0.0994) & 24139 (0.0000) && 24160 (0.0870) & 24157 (0.0746) && 24139 \\
                X-n469-k138 && 222152.2 (0.1480) & 222062 (0.1073) && 222158.6 (0.1508) & 222093 (0.1213) && 221824 \\
                X-n480-k70 && 89506.8 (0.0646) & 89499 (0.0559) && 89515 (0.0738) & 89491 (0.0470) && 89449 \\
                X-n491-k59 && 66654.4 (0.2518) & 66626 (0.2091) && 66633.8 (0.2208) & 66590 (0.1549) && 66487 \\
                X-n502-k39 && 69248.2 (0.0321) & 69234 (0.0116) && 69256.4 (0.0439) & 69238 (0.0173) && 69226 \\
                X-n513-k21 && 24201 (0.0000) & 24201 (0.0000) && 24201 (0.0000) & 24201 (0.0000) && 24201 \\
                X-n524-k153 && 154750.2 (0.1017) & 154648 (0.0356) && 154652.2 (0.0383) & 154610 (0.0110) && 154593 \\
                X-n536-k96 && 95063.4 (0.2060) & 95042 (0.1834) && 95084 (0.2277) & 95001 (0.1402) && 94868 \\
                X-n548-k50 && 86753.2 (0.0614) & 86745 (0.0519) && 86745.4 (0.0524) & 86709 (0.0104) && 86700 \\
                X-n561-k42 && 42732.6 (0.0365) & 42719 (0.0047) && 42736.6 (0.0459) & 42719 (0.0047) && 42717 \\
                X-n573-k30 && 50823 (0.2960) & 50777 (0.2052) && 50780.8 (0.2127) & 50762 (0.1756) && 50673 \\
                X-n586-k159 && 190582.8 (0.1402) & 190475 (0.0835) && 190541.2 (0.1183) & 190449 (0.0699) && 190316 \\
                X-n599-k92 && 108725.8 (0.2534) & 108663 (0.1955) && 108714.6 (0.2431) & 108581 (0.1199) && 108451 \\
                X-n613-k62 && 59711.6 (0.2966) & 59646 (0.1864) && 59682.8 (0.2483) & 59658 (0.2066) && 59535 \\
                X-n627-k43 && 62403 (0.3845) & 62354 (0.3056) && 62337.8 (0.2796) & 62284 (0.1930) && 62164 \\
                X-n641-k35 && 63891.6 (0.3102) & 63869 (0.2748) && 63858 (0.2575) & 63815 (0.1900) && 63694 \\
                X-n655-k131 && 106818.8 (0.0363) & 106810 (0.0281) && 106813 (0.0309) & 106809 (0.0272) && 106780 \\
                X-n670-k130 && 146836 (0.3444) & 146675 (0.2344) && 146815.8 (0.3306) & 146701 (0.2522) && 146332 \\
                X-n685-k75 && 68355.4 (0.2205) & 68284 (0.1158) && 68347 (0.2082) & 68292 (0.1276) && 68205 \\
                X-n701-k44 && 82339 (0.5078) & 82201 (0.3393) && 82169.8 (0.3013) & 82107 (0.2246) && 81923 \\
                X-n716-k35 && 43507.4 (0.2775) & 43446 (0.1360) && 43521.4 (0.3098) & 43494 (0.2466) && 43387 \\
                X-n733-k159 && 136445.6 (0.1877) & 136397 (0.1520) && 136436.4 (0.1809) & 136387 (0.1447) && 136190 \\
                X-n749-k98 && 77648.8 (0.4330) & 77600 (0.3699) && 77676.4 (0.4687) & 77597 (0.3660) && 77314 \\
                X-n766-k71 && 114773 (0.2787) & 114739 (0.2490) && 114715.8 (0.2287) & 114648 (0.1695) && 114454 \\
                X-n783-k48 && 72812.6 (0.5782) & 72763 (0.5097) && 72729.4 (0.4633) & 72637 (0.3357) && 72394 \\
                X-n801-k40 && 73500 (0.2660) & 73431 (0.1719) && 73501.4 (0.2679) & 73462 (0.2142) && 73305 \\
                X-n819-k171 && 158572.4 (0.2855) & 158511 (0.2466) && 158505.2 (0.2430) & 158439 (0.2011) && 158121 \\
                X-n837-k142 && 194352.4 (0.3176) & 194261 (0.2705) && 194237.4 (0.2583) & 194035 (0.1538) && 193737 \\
                X-n856-k95 && 89028.2 (0.0710) & 88990 (0.0281) && 89011.2 (0.0519) & 88993 (0.0315) && 88965 \\
                X-n876-k59 && 99681 (0.3847) & 99576 (0.2790) && 99715.6 (0.4195) & 99602 (0.3051) && 99299 \\
                X-n895-k37 && 54140.8 (0.5214) & 54063 (0.3769) && 54120.2 (0.4831) & 54101 (0.4475) && 53860 \\
                X-n916-k207 && 329930.2 (0.2282) & 329667 (0.1482) && 329879.8 (0.2129) & 329658 (0.1455) && 329179 \\
                X-n936-k151 && 133508.8 (0.5905) & 133314 (0.4438) && 133424.8 (0.5273) & 133176 (0.3398) && 132725 \\
                X-n957-k87 && 85556.8 (0.1074) & 85514 (0.0573) && 85581.6 (0.1364) & 85540 (0.0878) && 85465 \\
                X-n979-k58 && 119407 (0.3530) & 119326 (0.2849) && 119292.8 (0.2570) & 119257 (0.2269) && 118987 \\
                X-n1001-k43 && 72701.4 (0.4732) & 72631 (0.3759) && 72710.8 (0.4862) & 72620 (0.3607) && 72359 \\
                \midrule
                    Average Gap &&0.120 & && 0.105 & && \\
                    Median Gap &&0.059 & && 0.046 & && \\
                \bottomrule   
            \end{tabular}
        }
    \end{center}
\end{table}

\newpage

\begin{landscape} 
\subsection{Detailed Experiment using Baseline HGS-TW}
\label{sec:app-detailed-hgs-tw}

\subsubsection{Experiment on Clustered (C) Group: Set 1 and Set 2}
\label{subsec:app-tw-instance-c}

\begin{table}[H] 
    \begin{center}
        \caption{Performances variant $\mu$ on HGS-TW on clustered instances set 1 and set 2.}
        \label{table:generalized-tw-detailed-result-1}
        \vspace*{0.2cm}
        \setlength\tabcolsep{4pt}
        \scalebox{0.55}{ 
            \begin{tabular}{l rr rr r c|c l rr rr r}
                \toprule
                    \multirow{2}{*}{\textbf{Instance}} 
                    & \multicolumn{2}{c}{\textbf{HGS-TW}} 
                    & \multicolumn{2}{c}{\textbf{HGS-TW-$\mu$}} 
                    & \multirow{2}{*}{\textbf{BKS}} 
                    &&& \multirow{2}{*}{\textbf{Instance}} 
                    & \multicolumn{2}{c}{\textbf{HGS-TW}} 
                    & \multicolumn{2}{c}{\textbf{HGS-TW-$\mu$}} 
                    & \multirow{2}{*}{\textbf{BKS}}  \\
                    \cmidrule(lr){2-3} \cmidrule(lr){4-5} \cmidrule(lr){10-11} \cmidrule(lr){12-13}
                    & \textbf{Avg Gap} & \textbf{Best Gap} 
                    & \textbf{Avg Gap} & \textbf{Best Gap} 
                    &&&& 
                    & \textbf{Avg Gap} & \textbf{Best Gap} 
                    & \textbf{Avg Gap} & \textbf{Best Gap} 
                    & \\
                \midrule
                    C1\_8\_1 & 11632.68 (0.007) & 11631.9 (0.000) & 11633.04 (0.010) & 11631.9 (0.000) & 11631.9 &&& C2\_8\_1 & 11632.68 (0.007) & 11631.9 (0.000) & 11633.04 (0.010) & 11631.9 (0.000) & 11631.9 \\
                    C1\_8\_1 & 25156.9 (0.000) & 25156.9 (0.000) & 25156.90 (0.000) & 25156.9 (0.000) & 25156.9 &&& C2\_8\_1 & 11632.68 (0.007) & 11631.9 (0.000) & 11633.04 (0.010) & 11631.9 (0.000) & 11631.9 \\
                    C1\_8\_2 & 24995.76 (0.087) & 24978.9 (0.019) & 24999.42 (0.101) & 24983.7 (0.038) & 24974.1 &&& C2\_8\_2 & 11395.24 (0.006) & 11394.6 (0.001) & 11394.86 (0.003) & 11394.5 (0.000) & 11394.5 \\
                    C1\_8\_3 & 24226.98 (0.293) & 24205.4 (0.204) & 24219.80 (0.264) & 24185.3 (0.121) & 24156.1 &&& C2\_8\_3 & 11138.16 (0.001) & 11138.1 (0.000) & 11138.24 (0.001) & 11138.2 (0.001) & 11138.1 \\
                    C1\_8\_4 & 23855.1 (0.243) & 23831.2 (0.142) & 23861.74 (0.271) & 23850.3 (0.223) & 23797.3 &&& C2\_8\_4 & 10662.26 (0.213) & 10653.6 (0.132) & 10659.46 (0.187) & 10647.9 (0.078) & 10639.6 \\
                    C1\_8\_5 & 25138.6 (0.000) & 25138.6 (0.000) & 25138.60 (0.000) & 25138.6 (0.000) & 25138.6 &&& C2\_8\_5 & 11395.76 (0.001) & 11395.6 (0.000) & 11395.60 (0.000) & 11395.6 (0.000) & 11395.6 \\
                    C1\_8\_6 & 25133.3 (0.000) & 25133.3 (0.000) & 25133.30 (0.000) & 25133.3 (0.000) & 25133.3 &&& C2\_8\_6 & 11318.84 (0.022) & 11318.2 (0.017) & 11319.54 (0.029) & 11318.2 (0.017) & 11316.3 \\
                    C1\_8\_7 & 25127.3 (0.000) & 25127.3 (0.000) & 25127.30 (0.000) & 25127.3 (0.000) & 25127.3 &&& C2\_8\_7 & 11334.88 (0.017) & 11332.9 (0.000) & 11334.36 (0.013) & 11332.9 (0.000) & 11332.9 \\
                    C1\_8\_8 & 24892.3 (0.333) & 24867.7 (0.234) & 24880.54 (0.286) & 24861.0 (0.207) & 24809.7 &&& C2\_8\_8 & 11136.78 (0.026) & 11133.9 (0.000) & 11139.28 (0.048) & 11135.0 (0.010) & 11133.9 \\
                    C1\_8\_9 & 24275.18 (0.309) & 24253 (0.217) & 24282.06 (0.337) & 24267.2 (0.276) & 24200.4 &&& C2\_8\_9 & 11141.84 (0.013) & 11140.4 (0.000) & 11141.50 (0.010) & 11140.4 (0.000) & 11140.4 \\
                    C1\_8\_10 & 24114.12 (0.364) & 24083.1 (0.235) & 24081.06 (0.226) & 24059.7 (0.137) & 24026.7 &&& C2\_8\_10 & 10947.98 (0.018) & 10946.6 (0.005) & 10948.78 (0.025) & 10948.4 (0.022) & 10946 \\
                    C1\_10\_1 & 42444.8 (0.000) & 42444.8 (0.000) & 42444.80 (0.000) & 42444.8 (0.000) & 42444.8 &&& C2\_10\_1 & 16841.1 (0.000) & 16841.1 (0.000) & 16841.10 (0.000) & 16841.1 (0.000) & 16841.1 \\
                    C1\_10\_2 & 41450.68 (0.273) & 41390 (0.126) & 41434.02 (0.233) & 41393.2 (0.134) & 41337.8 &&& C2\_10\_2 & 16469.26 (0.040) & 16466.5 (0.024) & 16468.30 (0.035) & 16466.5 (0.024) & 16462.6 \\
                    C1\_10\_3 & 40327.56 (0.668) & 40159.2 (0.248) & 40271.06 (0.527) & 40219.2 (0.397) & 40060 &&& C2\_10\_3 & 16042.26 (0.036) & 16038 (0.009) & 16039.16 (0.017) & 16037.5 (0.006) & 16036.5 \\
                    C1\_10\_4 & 39618.56 (0.468) & 39535.8 (0.258) & 39582.70 (0.377) & 39539.0 (0.266) & 39434.1 &&& C2\_10\_4 & 15484.64 (0.163) & 15463.2 (0.024) & 15477.56 (0.117) & 15462.4 (0.019) & 15459.5 \\
                    C1\_10\_5 & 42434.8 (0.000) & 42434.8 (0.000) & 42434.80 (0.000) & 42434.8 (0.000) & 42434.8 &&& C2\_10\_5 & 16536.88 (0.094) & 16535.2 (0.084) & 16524.88 (0.022) & 16523.4 (0.013) & 16521.3 \\
                    C1\_10\_6 & 42437 (0.000) & 42437 (0.000) & 42437.00 (0.000) & 42437.0 (0.000) & 42437 &&& C2\_10\_6 & 16300.58 (0.061) & 16294 (0.020) & 16298.60 (0.048) & 16295.9 (0.032) & 16290.7 \\
                    C1\_10\_7 & 42421.5 (0.003) & 42420.7 (0.001) & 42420.72 (0.001) & 42420.4 (0.000) & 42420.4 &&& C2\_10\_7 & 16381.48 (0.019) & 16380.1 (0.010) & 16379.84 (0.009) & 16379.2 (0.005) & 16378.4 \\
                    C1\_10\_8 & 42082.86 (1.044) & 42026.5 (0.909) & 42013.94 (0.879) & 41881.4 (0.560) & 41648 &&& C2\_10\_8 & 16037.54 (0.053) & 16030.5 (0.009) & 16031.98 (0.018) & 16029.8 (0.004) & 16029.1 \\
                    C1\_10\_9 & 40822.48 (1.326) & 40621.7 (0.827) & 40735.96 (1.111) & 40575.3 (0.712) & 40288.4 &&& C2\_10\_9 & 16084.4 (0.056) & 16081.6 (0.039) & 16087.72 (0.077) & 16080.1 (0.029) & 16075.4 \\
                    C1\_10\_10 & 40228.02 (1.033) & 40001.2 (0.463) & 40133.58 (0.796) & 40003.1 (0.468) & 39816.8 &&& C2\_10\_10 & 15741.88 (0.084) & 15728.6 (0.000) & 15737.96 (0.060) & 15732.3 (0.024) & 15728.6 \\
                \midrule
                    Average Gap & 0.322 & & 0.270 & &&&& & 0.047 & & 0.036 & & \\
                    Median Gap & 0.258 & & 0.230 & &&&& & 0.024 & & 0.020 & & \\
                \bottomrule   
            \end{tabular}
        }
    \end{center}
\end{table}

\newpage

\subsubsection{Experiment on Random (R) Group: Set 1 and Set 2}
\label{subsec:app-tw-instance-r}

\begin{table}[H] 
    \begin{center}
        \caption{Performances variant $\mu$ on HGS-TW on random distribution instances set 1 and set 2.}
        \label{table:generalized-tw-detailed-result-2}
        \vspace*{0.2cm}
        \setlength\tabcolsep{4pt}
        \scalebox{0.55}{ 
            \begin{tabular}{l rr rr r c|c l rr rr r}
                \toprule
                    \multirow{2}{*}{\textbf{Instance}} 
                    & \multicolumn{2}{c}{\textbf{HGS-TW}} 
                    & \multicolumn{2}{c}{\textbf{HGS-TW-$\mu$}} 
                    & \multirow{2}{*}{\textbf{BKS}} 
                    &&& \multirow{2}{*}{\textbf{Instance}} 
                    & \multicolumn{2}{c}{\textbf{HGS-TW}} 
                    & \multicolumn{2}{c}{\textbf{HGS-TW-$\mu$}} 
                    & \multirow{2}{*}{\textbf{BKS}}  \\
                    \cmidrule(lr){2-3} \cmidrule(lr){4-5} \cmidrule(lr){10-11} \cmidrule(lr){12-13}
                    & \textbf{Avg Gap} & \textbf{Best Gap} 
                    & \textbf{Avg Gap} & \textbf{Best Gap} 
                    &&&& 
                    & \textbf{Avg Gap} & \textbf{Best Gap} 
                    & \textbf{Avg Gap} & \textbf{Best Gap} 
                    & \\
                \midrule
                    R1\_8\_1 & 36537.54 (0.530) & 36458.8 (0.313) & 36541.32 (0.540) & 36477.4 (0.364) & 36345 &&& R2\_8\_1 & 25061.74 (0.392) & 25022 (0.233) & 11676.26 (0.381) & 11664.7 (0.282) & 24963.8 \\
                    R1\_8\_2 & 32731.3 (1.406) & 32634.6 (1.106) & 32615.62 (1.047) & 32522.9 (0.760) & 32277.6 &&& R2\_8\_2 & 21395.4 (0.390) & 21341.6 (0.138) & 11426.99 (0.285) & 11411.1 (0.146) & 21312.2 \\
                    R1\_8\_3 & 29667.04 (1.249) & 29641.6 (1.162) & 29593.06 (0.996) & 29567.2 (0.908) & 29301.2 &&& R2\_8\_3 & 17282.72 (0.309) & 17259.9 (0.176) & 11162.72 (0.221) & 11151.0 (0.116) & 17229.5 \\
                    R1\_8\_4 & 27978.94 (0.881) & 27900.6 (0.598) & 27941.20 (0.745) & 27910.2 (0.633) & 27734.7 &&& R2\_8\_4 & 13269.88 (0.890) & 13218.8 (0.502) & 10714.30 (0.702) & 10681.2 (0.391) & 13152.8 \\
                    R1\_8\_5 & 33729.92 (0.704) & 33658.2 (0.490) & 33716.98 (0.666) & 33650.8 (0.468) & 33494 &&& R2\_8\_5 & 22870.82 (0.330) & 22854.6 (0.259) & 11436.54 (0.359) & 11416.7 (0.185) & 22795.6 \\
                    R1\_8\_6 & 31141.02 (0.870) & 31076.8 (0.662) & 31185.88 (1.015) & 31081.0 (0.676) & 30872.4 &&& R2\_8\_6 & 19838.28 (0.495) & 19786.7 (0.234) & 11363.46 (0.417) & 11334.1 (0.158) & 19740.5 \\
                    R1\_8\_7 & 29075.22 (0.994) & 28979 (0.660) & 29023.52 (0.815) & 28972.4 (0.637) & 28789 &&& R2\_8\_7 & 16437.22 (0.526) & 16411.2 (0.367) & 11396.33 (0.560) & 11381.8 (0.431) & 16351.2 \\
                    R1\_8\_8 & 27802.58 (0.700) & 27731.8 (0.443) & 27786.16 (0.640) & 27760.0 (0.545) & 27609.4 &&& R2\_8\_8 & 12737.82 (1.001) & 12713.7 (0.810) & 11214.93 (0.728) & 11174.2 (0.362) & 12611.6 \\
                    R1\_8\_9 & 32485.7 (0.708) & 32445.2 (0.583) & 32513.78 (0.795) & 32456.1 (0.616) & 32257.3 &&& R2\_8\_9 & 21389.78 (0.503) & 21376.4 (0.440) & 11183.99 (0.391) & 11157.4 (0.153) & 21282.7 \\
                    R1\_8\_10 & 31232.42 (1.016) & 31165.5 (0.800) & 31231.76 (1.014) & 31168.0 (0.808) & 30918.3 &&& R2\_8\_10 & 20125.78 (0.809) & 20101.7 (0.689) & 11016.28 (0.642) & 11002.6 (0.517) & 19964.2 \\
                    R1\_10\_1 & 53523.46 (0.938) & 53443 (0.786) & 53473.72 (0.844) & 53409.9 (0.724) & 53026.1 &&& R2\_10\_1 & 37164.3 (0.768) & 37088.4 (0.562) & 16944.40 (0.613) & 16911.0 (0.415) & 36881 \\
                    R1\_10\_2 & 49001.84 (1.534) & 48863.2 (1.247) & 48894.04 (1.310) & 48766.5 (1.046) & 48261.6 &&& R2\_10\_2 & 31506.34 (0.846) & 31452.4 (0.674) & 16574.54 (0.680) & 16549.6 (0.528) & 31241.9 \\
                    R1\_10\_3 & 45305.54 (1.415) & 45111.9 (0.982) & 45177.24 (1.128) & 45049.3 (0.842) & 44673.3 &&& R2\_10\_3 & 24532.2 (0.546) & 24447.3 (0.198) & 16115.90 (0.495) & 16077.2 (0.254) & 24399 \\
                    R1\_10\_4 & 42902.98 (1.089) & 42839.7 (0.940) & 42811.76 (0.874) & 42759.7 (0.752) & 42440.7 &&& R2\_10\_4 & 18096.72 (1.602) & 18015 (1.143) & 15612.56 (0.990) & 15536.9 (0.501) & 17811.4 \\
                    R1\_10\_5 & 50882.16 (0.943) & 50698.1 (0.578) & 50798.06 (0.776) & 50714.2 (0.610) & 50406.7 &&& R2\_10\_5 & 34414.98 (0.827) & 34374 (0.707) & 16656.54 (0.819) & 16634.7 (0.686) & 34132.8 \\
                    R1\_10\_6 & 47700.96 (1.647) & 47471.1 (1.157) & 47417.80 (1.043) & 47268.2 (0.725) & 46928.2 &&& R2\_10\_6 & 29408.2 (0.973) & 29270.5 (0.501) & 16425.70 (0.829) & 16413.8 (0.755) & 29124.7 \\
                    R1\_10\_7 & 44616.16 (1.406) & 44516.6 (1.180) & 44488.92 (1.117) & 44257.9 (0.592) & 43997.4 &&& R2\_10\_7 & 23298.1 (0.848) & 23222.3 (0.520) & 16533.16 (0.945) & 16444.8 (0.406) & 23102.2 \\
                    R1\_10\_8 & 42683.66 (0.956) & 42657.2 (0.894) & 42672.96 (0.931) & 42556.1 (0.655) & 42279.3 &&& R2\_10\_8 & 17601.84 (1.138) & 17557.7 (0.885) & 16155.48 (0.788) & 16109.8 (0.503) & 17403.7 \\
                    R1\_10\_9 & 49736.6 (1.167) & 49589.3 (0.868) & 49530.86 (0.749) & 49390.7 (0.464) & 49162.8 &&& R2\_10\_9 & 32288.76 (0.932) & 32224.3 (0.731) & 16200.53 (0.778) & 16180.7 (0.655) & 31990.6 \\
                    R1\_10\_10 & 48043.28 (1.433) & 47954.5 (1.245) & 47850.38 (1.026) & 47761.0 (0.837) & 47364.6 &&& R2\_10\_10 & 30195.06 (1.188) & 30063.2 (0.746) & 15868.03 (0.886) & 15801.7 (0.464) & 29840.5 \\
                \midrule
                    Average Gap & 1.079 & & 0.904 & &&&& & 0.766 & & 0.626 & & \\
                    Median Gap & 1.005 & & 0.903 & &&&& & 0.818 & & 0.661 & & \\
                \bottomrule   
            \end{tabular}
        }
    \end{center}
\end{table}

\newpage

\subsubsection{Experiment on Random-Clustered (RC) Group: Set 1 and Set 2}
\label{subsec:app-tw-instance-rc}

\begin{table}[H] 
    \begin{center}
        \caption{Performances variant $\mu$ on HGS-TW on random-clustered instances set 1 and set 2.}
        \label{table:generalized-tw-detailed-result-3}
        \vspace*{0.2cm}
        \setlength\tabcolsep{4pt}
        \scalebox{0.55}{ 
            \begin{tabular}{l rr rr r c|c l rr rr r}
                \toprule
                    \multirow{2}{*}{\textbf{Instance}} 
                    & \multicolumn{2}{c}{\textbf{HGS-TW}} 
                    & \multicolumn{2}{c}{\textbf{HGS-TW-$\mu$}} 
                    & \multirow{2}{*}{\textbf{BKS}} 
                    &&& \multirow{2}{*}{\textbf{Instance}} 
                    & \multicolumn{2}{c}{\textbf{HGS-TW}} 
                    & \multicolumn{2}{c}{\textbf{HGS-TW-$\mu$}} 
                    & \multirow{2}{*}{\textbf{BKS}}  \\
                    \cmidrule(lr){2-3} \cmidrule(lr){4-5} \cmidrule(lr){10-11} \cmidrule(lr){12-13}
                    & \textbf{Avg Gap} & \textbf{Best Gap} 
                    & \textbf{Avg Gap} & \textbf{Best Gap} 
                    &&&& 
                    & \textbf{Avg Gap} & \textbf{Best Gap} 
                    & \textbf{Avg Gap} & \textbf{Best Gap} 
                    & \\
                \midrule
                    RC1\_8\_1 & 30145.6 (0.644) & 30081.8 (0.431) & 30110.62 (0.527) & 30048.8 (0.321) & 29952.8 &&& RC2\_8\_1 & 19283.8 (0.430) & 19236.6 (0.184) & 30044.93 (0.308) & 30012.4 (0.199) & 19201.3 \\
                    RC1\_8\_2 & 28571.34 (0.994) & 28557.4 (0.945) & 28506.68 (0.766) & 28453.0 (0.576) & 28290.1 &&& RC2\_8\_2 & 16763.38 (0.322) & 16738.2 (0.172) & 28347.80 (0.204) & 28316.9 (0.095) & 16709.5 \\
                    RC1\_8\_3 & 27646.44 (0.724) & 27521 (0.267) & 27561.14 (0.413) & 27497.3 (0.181) & 27447.7 &&& RC2\_8\_3 & 14080.86 (0.480) & 14058.9 (0.323) & 27556.29 (0.396) & 27493.9 (0.168) & 14013.6 \\
                    RC1\_8\_4 & 26736.26 (0.674) & 26667.5 (0.415) & 26733.74 (0.665) & 26684.8 (0.480) & 26557.2 &&& RC2\_8\_4 & 11049.64 (0.731) & 11033.8 (0.587) & 26727.88 (0.643) & 26634.4 (0.291) & 10969.4 \\
                    RC1\_8\_5 & 29410.66 (0.653) & 29364.4 (0.495) & 29387.98 (0.575) & 29352.3 (0.453) & 29219.9 &&& RC2\_8\_5 & 17550.06 (0.481) & 17519.1 (0.303) & 29319.74 (0.342) & 29304.4 (0.289) & 17466.1 \\
                    RC1\_8\_6 & 29386.02 (0.814) & 29339.4 (0.654) & 29325.52 (0.607) & 29266.1 (0.403) & 29148.7 &&& RC2\_8\_6 & 17276.54 (0.500) & 17237.1 (0.270) & 29266.61 (0.405) & 29219.7 (0.244) & 17190.6 \\
                    RC1\_8\_7 & 28997.08 (0.916) & 28974.6 (0.837) & 28924.06 (0.661) & 28891.2 (0.547) & 28734 &&& RC2\_8\_7 & 16500.96 (0.848) & 16455.7 (0.571) & 29010.38 (0.962) & 28977.2 (0.846) & 16362.2 \\
                    RC1\_8\_8 & 28624.86 (0.827) & 28546.8 (0.552) & 28632.54 (0.854) & 28610.2 (0.776) & 28390 &&& RC2\_8\_8 & 15673.4 (0.931) & 15628 (0.639) & 28646.94 (0.905) & 28599.0 (0.736) & 15528.8 \\
                    RC1\_8\_9 & 28576.9 (0.866) & 28435.8 (0.368) & 28517.54 (0.656) & 28383.5 (0.183) & 28331.6 &&& RC2\_8\_9 & 15267.8 (0.597) & 15232.1 (0.362) & 28409.82 (0.276) & 28390.4 (0.208) & 15177.2 \\
                    RC1\_8\_10 & 28317.34 (0.528) & 28214.2 (0.162) & 28286.58 (0.419) & 28269.7 (0.359) & 28168.5 &&& RC2\_8\_10 & 14502.68 (0.917) & 14484.2 (0.788) & 28407.40 (0.848) & 28344.5 (0.625) & 14370.9 \\
                    RC1\_10\_1 & 46288.88 (1.088) & 46175.5 (0.840) & 46203.32 (0.901) & 46124.2 (0.728) & 45790.7 &&& RC2\_10\_1 & 28407.04 (1.011) & 28293.5 (0.608) & 46202.91 (0.900) & 46063.9 (0.597) & 28122.6 \\
                    RC1\_10\_2 & 44184.9 (1.160) & 44135.7 (1.047) & 44168.72 (1.123) & 44076.4 (0.911) & 43678.3 &&& RC2\_10\_2 & 24346.1 (0.402) & 24312.7 (0.264) & 43850.54 (0.394) & 43810.5 (0.303) & 24248.6 \\
                    RC1\_10\_3 & 42640.1 (1.230) & 42554.8 (1.028) & 42682.14 (1.330) & 42622.4 (1.188) & 42121.9 &&& RC2\_10\_3 & 19711.2 (0.475) & 19702.3 (0.429) & 42293.37 (0.407) & 42235.3 (0.269) & 19618.1 \\
                    RC1\_10\_4 & 41709.14 (0.850) & 41636.4 (0.675) & 41703.30 (0.836) & 41644.8 (0.695) & 41357.4 &&& RC2\_10\_4 & 15818.48 (1.046) & 15771.8 (0.748) & 41797.16 (1.063) & 41737.0 (0.918) & 15654.7 \\
                    RC1\_10\_5 & 45508.38 (1.067) & 45310.1 (0.626) & 45441.56 (0.918) & 45395.9 (0.817) & 45028.1 &&& RC2\_10\_5 & 26030.96 (0.905) & 25933.8 (0.528) & 45375.23 (0.771) & 45305.3 (0.616) & 25797.5 \\
                    RC1\_10\_6 & 45414.38 (1.150) & 45285.5 (0.863) & 45317.52 (0.934) & 45192.3 (0.655) & 44898.2 &&& RC2\_10\_6 & 26082.16 (1.162) & 25995.8 (0.827) & 45441.04 (1.209) & 45371.7 (1.055) & 25782.5 \\
                    RC1\_10\_7 & 44919.54 (1.150) & 44859.5 (1.014) & 44827.86 (0.943) & 44738.8 (0.743) & 44409 &&& RC2\_10\_7 & 24586.98 (0.802) & 24540.7 (0.612) & 44796.62 (0.873) & 44680.8 (0.612) & 24391.4 \\
                    RC1\_10\_8 & 44417.98 (1.142) & 44360.2 (1.010) & 44311.70 (0.900) & 44267.1 (0.798) & 43916.5 &&& RC2\_10\_8 & 23558.98 (1.199) & 23462.5 (0.785) & 44358.84 (1.007) & 44222.1 (0.696) & 23279.8 \\
                    RC1\_10\_9 & 44291.62 (0.989) & 44225.9 (0.839) & 44300.28 (1.008) & 44215.9 (0.816) & 43858 &&& RC2\_10\_9 & 22932.06 (0.882) & 22883.7 (0.669) & 44300.52 (1.009) & 44215.1 (0.814) & 22731.6 \\
                    RC1\_10\_10 & 43907.32 (0.858) & 43854.3 (0.736) & 43963.24 (0.987) & 43884.5 (0.806) & 43533.7 &&& RC2\_10\_10 & 21992.64 (1.203) & 21932.4 (0.926) & 43982.60 (1.031) & 43833.4 (0.688) & 21731.2 \\
                \midrule
                    Average Gap & 0.916 & & 0.801 & &&&& & 0.766 & & 0.698 & & \\
                    Median Gap & 0.891 & & 0.845 & &&&& & 0.825 & & 0.810 & & \\
                \bottomrule   
            \end{tabular}
        }
    \end{center}
\end{table}

\end{landscape}

\clearpage
\newpage
\printbibliography

\end{document}